\documentclass[letterpaper]{article} 
\usepackage{aaai2026}  
\usepackage{times}  
\usepackage{helvet}  
\usepackage{courier}  
\usepackage[hyphens]{url}  
\usepackage{graphicx} 
\usepackage{natbib}  
\usepackage{caption} 
\usepackage{algorithm}
\usepackage{algorithmic}

\usepackage{booktabs}
\usepackage{newfloat}
\usepackage{listings}
\usepackage{xcolor}         
\usepackage{amsmath}
\usepackage{subcaption}
\usepackage{amssymb}

\DeclareCaptionStyle{ruled}{labelfont=normalfont,labelsep=colon,strut=off} 
\floatstyle{ruled}
\newfloat{listing}{tb}{lst}{}
\floatname{listing}{Listing}
\title{How Does Alignment Enhance LLMs' Multilingual Capabilities? 

A Language Neurons Perspective}
\author{
\textbf{Shimao Zhang\textsuperscript{1}\thanks{Equal contribution.}\thanks{Work done during his internship at MSRA.}, }
\textbf{Zhejian Lai\textsuperscript{1}\footnotemark[1], }
\textbf{Xiang Liu\textsuperscript{1}\footnotemark[1], }
\textbf{Shuaijie She\textsuperscript{1}, }
\textbf{Xiao Liu\textsuperscript{2}, }\\
\textbf{Yeyun Gong\textsuperscript{2}, }
\textbf{Shujian Huang\textsuperscript{1}\thanks{Corresponding author.}, } 
\textbf{Jiajun Chen\textsuperscript{1}}\\
}
\affiliations{
\textsuperscript{1} National Key Laboratory for Novel Software Technology, Nanjing University \\
\textsuperscript{2} Microsoft Research Asia \\
\text{\{smzhang, laizj, liuxiang, shesj\}@smail.nju.edu.cn, \{huangsj, chenjj\}@nju.edu.cn},\\
\text{\{xiao.liu.msrasia, yegong\}@microsoft.com} \\
}

\usepackage{bibentry}

\begin{document}

\maketitle

\begin{abstract}

Multilingual Alignment is an effective and representative paradigm to enhance LLMs' multilingual capabilities, which transfers the capabilities from the high-resource languages to the low-resource languages. Meanwhile, some research on language-specific neurons provides a new perspective to analyze and understand LLMs' mechanisms. However, we find that there are many neurons that are shared by multiple but not all languages and cannot be correctly classified. In this work, we propose a ternary classification methodology that categorizes neurons into three types, including \textit{language-specific} neurons, \textit{language-related} neurons, and \textit{general} neurons. And we propose a corresponding identification algorithm to distinguish these different types of neurons. Furthermore, based on the distributional characteristics of different types of neurons, we divide the LLMs' internal process for multilingual inference into four parts: (1) multilingual understanding, (2) shared semantic space reasoning, (3) multilingual output space transformation, and (4) vocabulary space outputting. Additionally, we systematically analyze the models before and after alignment with a focus on different types of neurons. We also analyze the phenomenon of ``Spontaneous Multilingual Alignment''. Overall, our work conducts a comprehensive investigation based on different types of neurons, providing empirical results and valuable insights to better understand multilingual alignment and multilingual capabilities of LLMs.

\begin{links}
    \link{Code}{https://github.com/NJUNLP/Language-Neurons-Alignment}
    \link{Extended version}{https://arxiv.org/abs/2505.21505}
\end{links}
\end{abstract}

\section{Introduction}\label{sec:introduction}
By training on the extensive corpus, large language models~(LLMs) demonstrate outstanding language capabilities~\citep{yang2024qwen2,liu2024deepseek,grattafiori2024llama,zhang-etal-2025-process}. However, due to the unbalanced pretraining corpus across different languages, LLMs have very uneven performance on high-resource languages and low-resource languages~\citep{huang2023not,zhu2023multilingual,zhang-etal-2024-getting,fan2025slam}. Therefore, researchers have conducted comprehensive explorations to further enhance the multilingual performance of LLMs. A straightforward approach is increasing the proportion of non-English texts during pretraining~\citep{ni2021m3p,yang2024qwen2} or performing continual pretraining with multilingual texts~\citep{liu2021continual,ji2024emma}. But these approaches often entail high computational costs and substantial amounts of multilingual data.

\begin{figure}[tbp]
    \centering
    \includegraphics[width=0.65\linewidth]{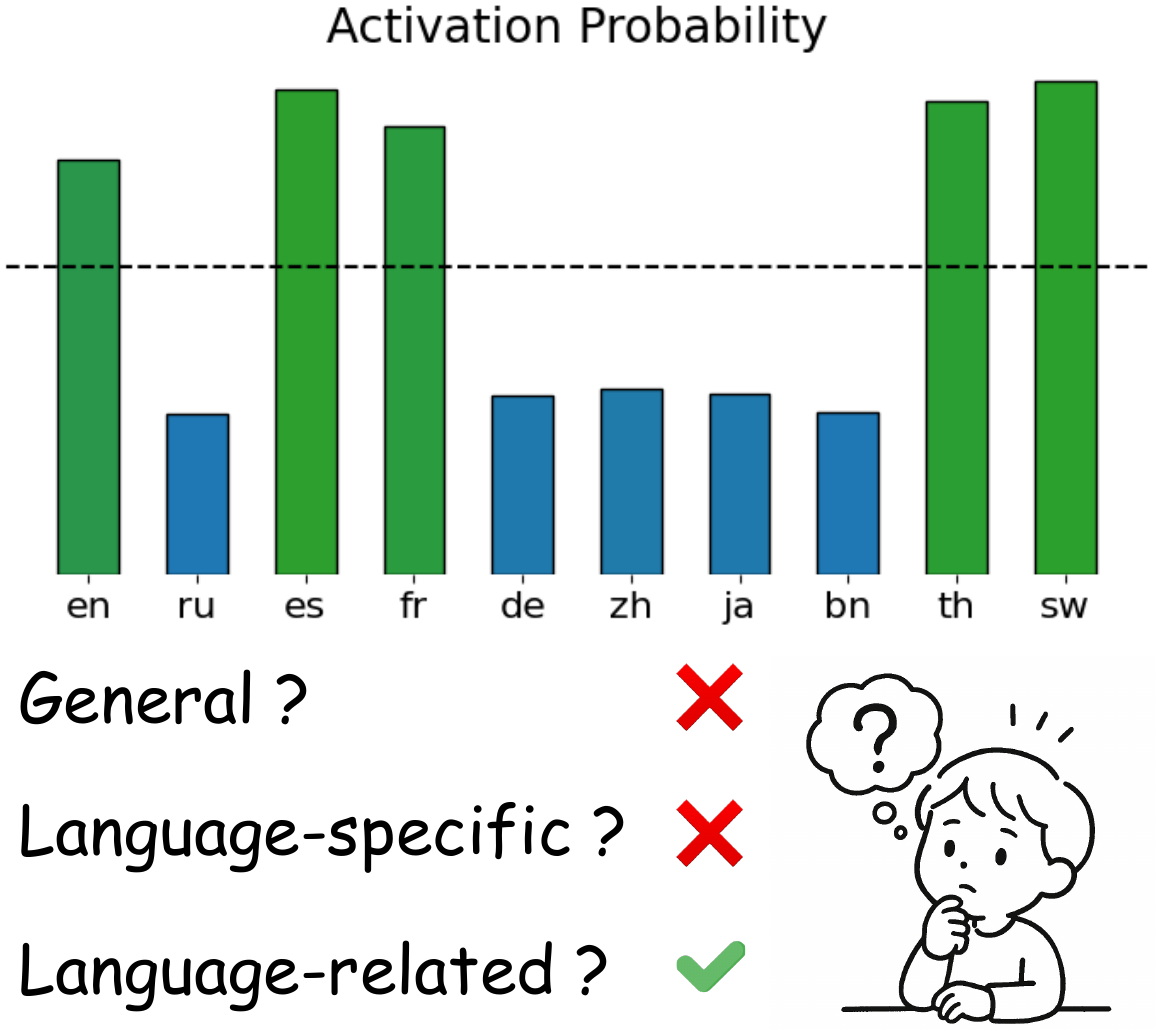}
    \caption{A neuron's activation probability across different languages. This neuron that exhibits high activation probabilities across \textit{multiple} (\textbf{green}) but \textit{not all} (\textbf{blue}) languages can't be correctly categorized under the existing classification methodology. The dashed line denotes the threshold that determines whether a neuron exhibits high activation in a given language.}
    \label{fig:case-study-language-related-neurons}
\end{figure}

Considering LLMs' great performance on high-resource languages, multilingual alignment has emerged as a representative paradigm for enhancing multilingual reasoning by transferring knowledge from high-resource to low-resource languages~\citep{zhao2024llama,she-etal-2024-mapo}. A representative example is MAPO~(Multilingual-Alignment-as-Preference Optimization)~\citep{she-etal-2024-mapo}, which improves multilingual alignment by utilizing a well-trained multilingual translation model to compute alignment scores based on the conditional generation probability of translating non-English responses into English.


Many studies conduct systematic mechanism analyses of the multilingual alignment and LLMs' multilingual capabilities. \citet{zhao2024large} split the multilingual processing workflow into three parts: multilingual understanding, resolving tasks, and generating outputs in the target language. This three-stage inference workflow clearly demonstrates how LLMs leverage English as a pivot language to handle multilingualism using a unified pattern.

Inspired by the neurobiological underpinnings of human language faculties,~\citet{tang-etal-2024-language} categorize neurons in LLMs into two primary types: language-specific neurons and general neurons. Notably, these language-specific neurons are primarily situated in the model's top and bottom layers~\citep{tang-etal-2024-language}, which is consistent with the three-stage multilingual workflow of~\citet{zhao2024large}. 


However, we identify a key limitation in the existing neuron classification methodology: the identification of language-specific neurons focuses on the processing a particular language and thus neglects the inter-language alignment; while general neurons encode universal knowledge which is independent of a specific language. This leads to a critical question: what is the mechanism of sharing neurons between some languages?
Furthermore, it is not well understood how multilingual alignment enhances the LLMs' multilingual reasoning capabilities from the perspective of language neurons.

In this work, we comprehensively investigate the multilingual alignment of LLMs with MAPO~\citep{she-etal-2024-mapo} as a representative multilingual alignment algorithm. As shown in Figure~\ref{fig:case-study-language-related-neurons}, we observed the existence of neurons that exhibit high activation probability on multiple but not all languages. These neurons are neither language-specific nor general for all languages. Hence, we refer to them as \textbf{language-related neurons}.



Moreover, we redefine \textbf{language-specific neurons} to restrict each neuron to be activated for only one language. The \textbf{general neurons} are defined as the neurons that are effective for all languages. To facilitate subsequent analysis, we propose a corresponding identification algorithm for distinguishing these different types of neurons.


Then we analyze the models before and after alignment, focusing on the changes in different types of neurons. Based on the distributional characteristics of these neurons, we divide LLMs' internal process for multilingual inference into four parts, with different parts exhibit distinct dependencies on different types of neurons. We demonstrate that multilingual alignment significantly enhances the activation of the corresponding types of neurons across the relevant layers. 
Additionally, we analyze the ``spontaneous multilingual alignment''~\citep{zhang-etal-2024-getting} phenomenon in LLMs, providing insights into the roles of general neurons and language-related neurons shared across languages. For further analysis, we also provide observations about the uniqueness of English and the neuron distributions. Overall, based on different types of neurons, we present empirical results and valuable insights that contribute to a deeper understanding of multilingual alignment and the multilingual reasoning capabilities of LLMs.

\section{Related Work}\label{sec:related-work}

\subsection{Multilingual Alignment}\label{subsec:bg-multilingual-alignment}
Conducting pretraining or continual pretraining on the multilingual corpus is a straightforward and effective method to enhance LLMs' multilingual capabilities~\citep{ni2021m3p,ji2024emma}. However, these methods typically require substantial investments in time, data, and computational resources. Thus, many researchers perform multilingual alignment to improve LLMs' multilingual performance by transferring the capabilities from high-resource languages to low-resource languages~\citep{eronen2023zero,zhao2024adamergex,zhao2024llama,she-etal-2024-mapo}, which efficiently and effectively improves the model performance in low-resource language scenarios. Furthermore, \citet{zhang-etal-2024-getting} first finds the ``Spontaneous Multilingual Alignment'' phenomenon in LLMs, which demonstrates that conducting multilingual alignment based on a small number of languages effectively improves the alignment even between English and many languages unseen during alignment.

\subsection{Mechanistic Interpretability}\label{subsec:bg-mechanistic-interpretability}
In addition to enhancing LLMs' multilingual performance, research on the underlying mechanisms of multilingual capabilities in LLMs is still ongoing. It is crucial for us to understand and explain the LLMs and related methods explicitly. Typically, the existing approaches primarily perform mechanistic interpretability analyses by observing the internal states of the model~\citep{logit-lens,zhang-etal-2024-getting,zhao2024large,mousi-etal-2024-exploring}. Overall, neuron states and latent intermediate logits are both important objects of observation. For latent logits, \citet{wendler2024llamas} utilizes logit lens~\citep{logit-lens} to directly project the logits in the intermediate layers to the vocabulary space, which reveals the latent participation of English in the intermediate layers. For neuron states, \citet{hu2024large} analyzes the neuron activation overlap to measure the extent of shared neuron activation across different languages. 

\subsection{Language-Specific Neurons}\label{subsec:bg-language-specific-neurons}
Many studies have revealed the language-related and universal components in LLMs. At the layer level, the multilingual processing of LLMs is considered to involve three stages~\citep{zhao2024large,wendler2024llamas}: converting multilingual inputs into a shared semantic space, intermediate-layer reasoning, and outputting in the target language. The top and bottom layers of the model handle multilingual processing, while the intermediate layers perform inference in similar patterns across different languages. This demonstrates a distinct division of labor within the model at the layer level regarding language specificity.


Furthermore, many studies investigate the finer-grained methods for language-specific neuron identification~\citep{kojima2024multilingual,tang-etal-2024-language,tan2024neuron}. \citet{tang-etal-2024-language} categorizes activated neurons into language-specific neurons and general neurons. They detect language-specific neurons by calculating language activation probability entropy on massive text. However, we find that some neurons are activated by multiple languages (i.e., not language-specific), yet are not universally activated across all languages (i.e., not general). Simply categorizing activated neurons into two classes blurs this distinction. Thus, we propose a new method to identify neurons, which categorizes activated neurons into three types: language-specific neurons, language-related neurons, and general neurons.

\section{Methodology}\label{sec:methodology}


\subsection{Preliminary Study}\label{subsec:method-preliminary-study}
Most LLMs are pretrained mainly on the high-resource language corpus, which leads to LLMs' unstable and unbalanced performance in multilingual scenarios. As a representative multilingual alignment algorithm, Multilingual-Alignment-as-Preference Optimization~(MAPO)~\citep{she-etal-2024-mapo} effectively and efficiently improves the LLMs' multilingual performance. Additionally, it is also important for us to understand and analyze the mechanism of LLMs' multilingual capabilities and multilingual alignment. Moreover, some studies on the identification of the language-specific and general neurons in LLMs~\citep{tang-etal-2024-language,kojima2024multilingual}. It is found that LLMs' capabilities of processing a particular language mainly come from a small subset of neurons~\citep{tang-etal-2024-language}.


However, there are still many important questions waiting for further investigation. On the one hand, many methods overlook neurons activated by multiple languages yet not general, namely language-related neurons lie between language-specific and general categories. On the other hand, research from the perspective of language neurons on the underlying mechanisms of LLMs' multilingual alignment and multilingual reasoning capabilities remains quite limited, which is essential for better understanding and improving the multilingual performance of LLMs. 

\subsection{Multilingual Alignment}\label{subsec:method-multilingual-alignment}

MAPO is a typical multilingual alignment algorithm to align the reasoning capabilities of non-English language responses with those of English, which serves as the pivot language.
Specifically, for a given query $X$ in a target (non-English) language and its corresponding English variant $X_{En}$, we collect their respective responses $Y$ and $Y_{En}$. An off-the-shelf translation model, parameterized by $\theta$, is deployed to estimate the conditional generation probability $P(Y \mid Y_{En}; \theta)$ by force-decoding $Y$ conditioned on $Y_{En}$. A higher conditional probability is interpreted as stronger alignment between the target language response and its English counterpart. This probability is then used as an alignment score, denoted $r_{\theta}(X, Y)$.

This alignment score can be integrated into preference optimization algorithms. In PPO~\citep{ppo}, $r_{\theta}(X, Y)$ can be directly employed as the reward score. In DPO~\citep{dpo}, for each target language, $n$ distinct outputs are generated. Based on the alignment score, these $n$ outputs are used to form $\binom{n}{2}$ preference pairs $(Y_w, Y_l)$, where $Y_w$ is deemed superior to $Y_l$ due to a higher alignment score. The model is optimized by Eq.~\ref{eq:dpo-A} to Eq.~\ref{eq:dpo}:
\begin{align}
\label{eq:dpo-A}
A&=\beta\log \frac{\pi_{\theta}(Y_w | X)}{\pi_{ref}(Y_w | X)} \\
\label{eq:dpo-B}
B&=\beta\log \frac{\pi_{\theta}(Y_l | X)}{\pi_{ref}(Y_l | X)} \\
\label{eq:dpo}
L_{DPO}(\pi_{\theta}; \pi_{ref}) &= -E_{(X, Y_w, Y_l) \sim D} \left[ \log \sigma \left(  A -  B \right) \right] 
\end{align}


\begin{figure*}[tbp]
    \centering
    \begin{minipage}[b]{0.33\textwidth}
    \includegraphics[width=1.0\linewidth]{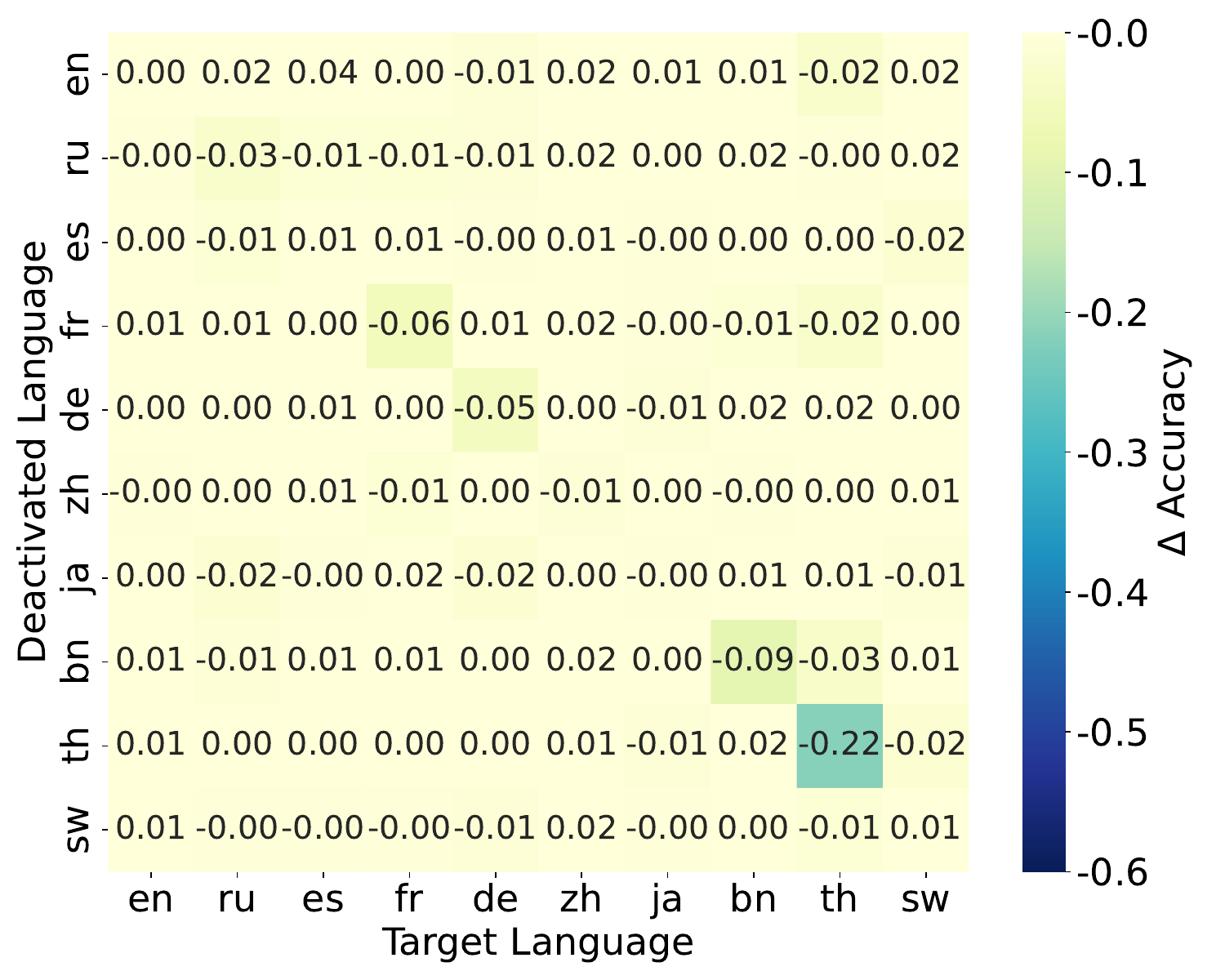}
    \subcaption{Specific Neurons}
    \label{fig:mistral-mgsm-base-language-specific-heatmap}
    \end{minipage}
    \begin{minipage}[b]{0.33\textwidth}
    \includegraphics[width=1.0\linewidth]{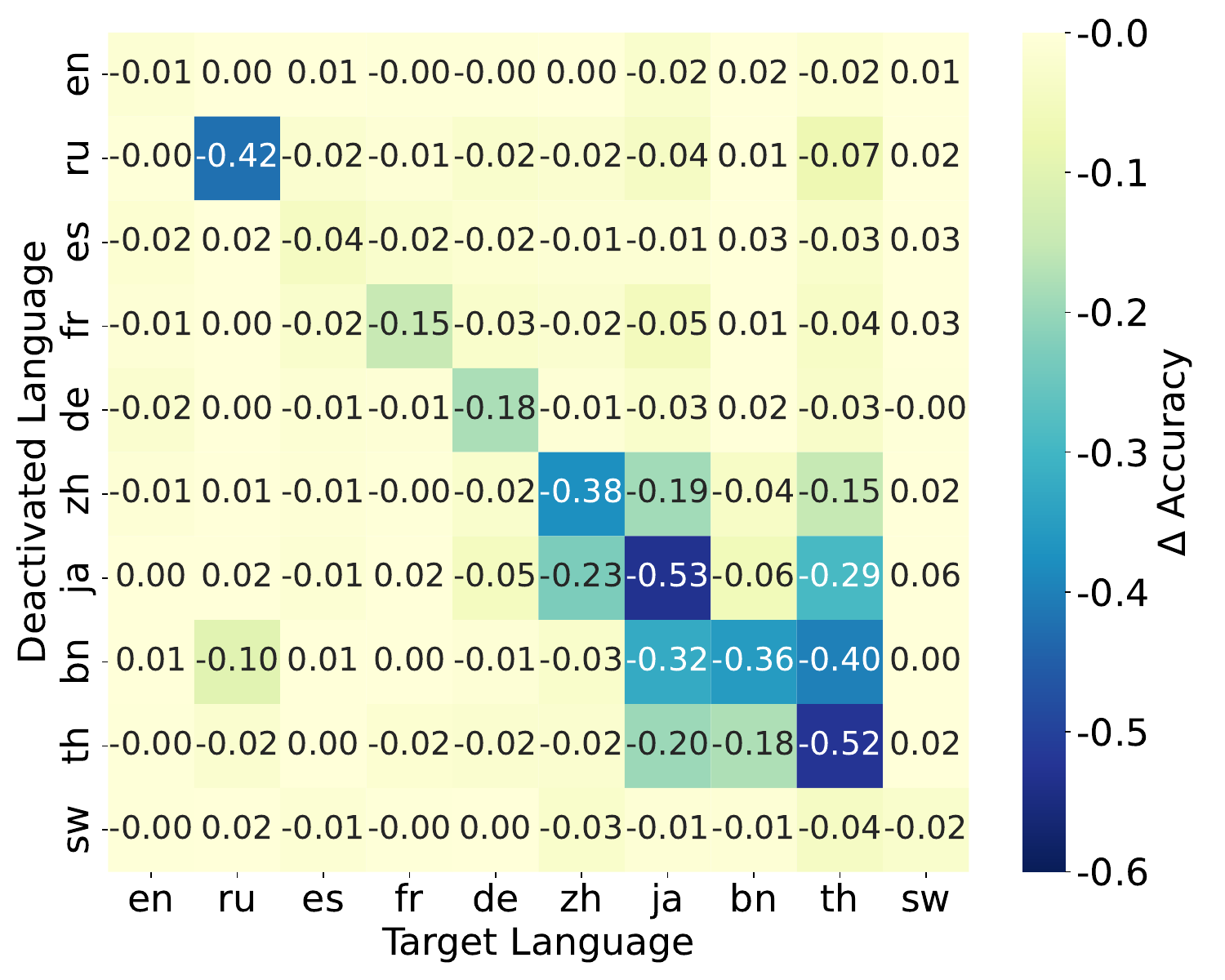}
    \subcaption{Specific \& Related Neurons}
    \label{fig:mistral-mgsm-base-language-heatmap}
    \end{minipage}
    \begin{minipage}[b]{0.33\textwidth}
    \includegraphics[width=1.0\linewidth]{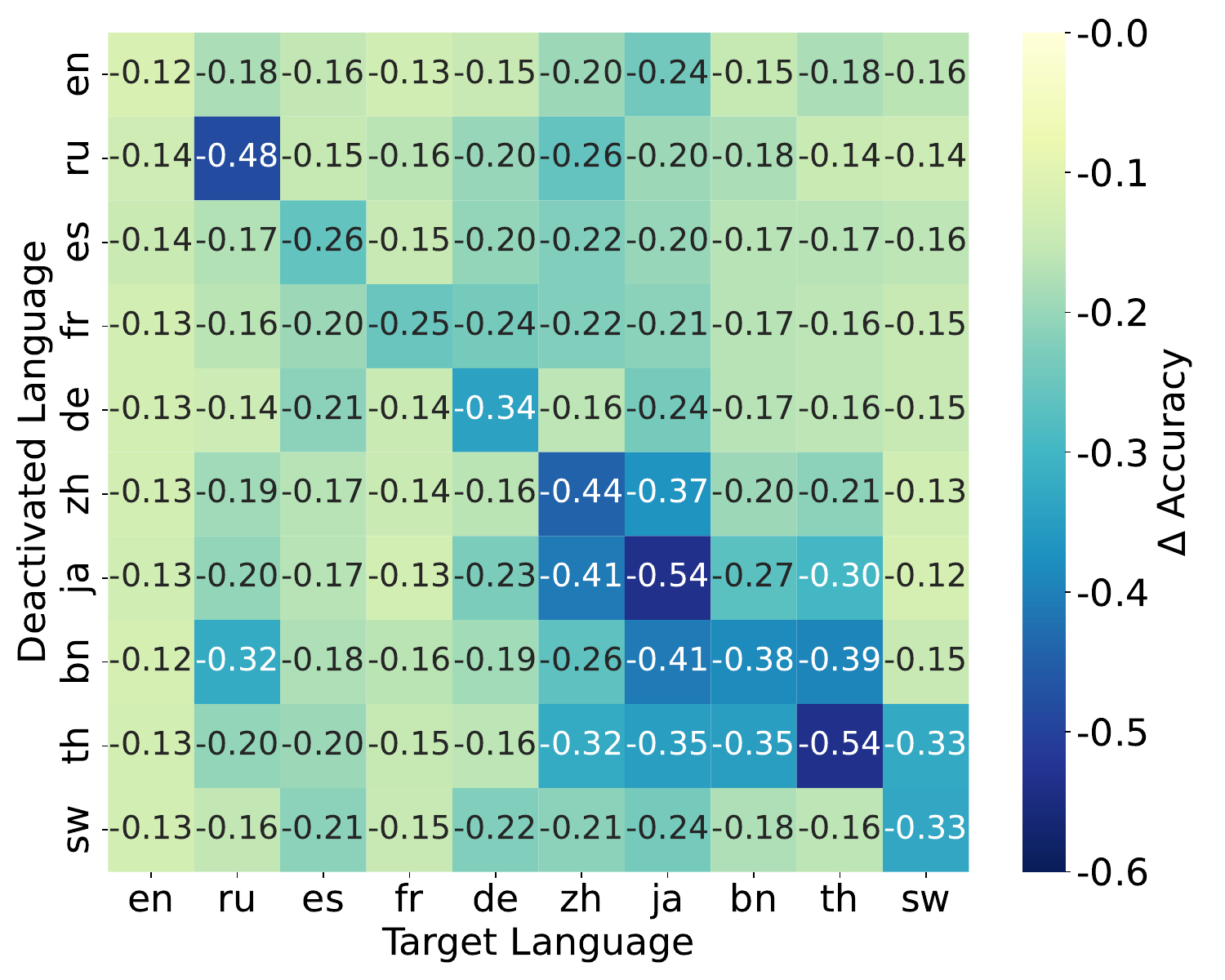}
    \subcaption{Specific \& Related \& General Neurons}
    \label{fig:mistral-mgsm-base-general-heatmap}
    \end{minipage}
\caption{Accuracy changes of MistralMathOctopus on MGSM after deactivating language-specific neurons or language-specific \& language-related neurons or language-specific \& language-related \& general neurons. For comparison, the results of~\citet{tang-etal-2024-language} are provided in appendix.}
\label{fig:mistral-mgsm-heatmap-acc}
\end{figure*}

\subsection{Language Neurons Identification}\label{subsec:method-language-neurons-detection}


Following \citet{tang-etal-2024-language}, the neurons in our work are defined as a linear transformation of a single column in a weighted matrix $W$ followed by a non-linear activation, SiLU~\citep{silu}. For the $j$-th neuron in the $i$-th layer, its activation probability when processing responses in language $k$ is computed as:
\begin{equation}
p_{i,j}^k=\mathbb{E}\left(\mathbb{I}\left(\mathrm{SiLU}(x^i W^i)_j > 0\right)\mid \text{language}\ k \right)
\end{equation}

We define language-related neurons as those neurons exhibiting high activation probabilities for multiple but not all languages. However, methods that only focus on language specificity are defective in detecting these neurons, while these neurons are important for multilingual reasoning but exhibit low language specificity. To better detect the different types of neurons, we trade off two intrinsic properties of each neuron: (1)~\textit{Language-specificity}, which is quantified by the entropy of its activation probability distribution across languages; (2)~\textit{Effectiveness}, which measures the extent to which the neuron participates when the model solves tasks. And the effectiveness is quantified by the neuron's maximum activation probability across different languages in our work. By doing so, we are better able to simultaneously identify neurons that are relatively more specific and relatively more generalized. They are combined into a unified metric as formulated in Eq~\ref{eq:metric}:
\begin{equation}
\label{eq:metric}
\mathrm{score}_{i,j} = -\sum_{k=1}^{l} p_{i,j}^{\prime k}\,\log p_{i,j}^{\prime k}\ -\ \lambda\,\max_{1\le k\le l} p_{i,j}^k,
\end{equation}


where $p_{i,j}^{\prime}$ represents the probability distribution $p_{i,j}$ after $\rm L1$ normalization and $\lambda$ is a balancing coefficient. Applying L1 normalization converts $p_{i,j}$~(raw probabilities on the target languages) into a valid probability distribution $p_{i,j}^{\prime}$~\citep{tang-etal-2024-language}. Specifically, we automatically determine a value of $\lambda$ such that general neurons are not identified as language-specific or language-related neurons shared by all $l$ languages. Following~\citet{tang-etal-2024-language}, neurons with scores falling in the lowest $1\%$ are selected.

Furthermore, to identify how many languages each selected neuron is related to, we introduce a threshold $\tau$:
\begin{equation}
N_{i,j} = \sum_{k=1}^l \mathbb{I}\bigl(p_{i,j}^k > \tau\bigr).
\end{equation}


Following~\citet{tang-etal-2024-language}, the threshold $\tau$ is set to the top 5\% of all activation probabilities. A neuron is considered as a \textbf{language-specific neuron} if $N_{i,j} = 1$, and as a \textbf{language-related neuron} if $1 < N_{i,j} < l$. Meanwhile, a neuron is considered \textbf{general neuron} if it exhibits high activation probabilities across all $l$ languages.

Finally, given our focus on multilingual reasoning tasks, we select neurons exclusively based on responses from multilingual reasoning datasets, rather than relying on multilingual plain text~\citep{tang-etal-2024-language}.

\section{Experiments}\label{sec:experiments}

\subsection{Experimental Setup}\label{subsec:experimental-setup}
Following~\citet{she-etal-2024-mapo}, we conduct our experiments and analyzes on the mathematical reasoning tasks in different languages. In this section, we introduce our experimental settings in detail.

\paragraph{Models} We include two different models in our experiments and analyses. Following~\citet{she-etal-2024-mapo}, we conduct our experiments on MistralMathOctopus-7B~\footnote{\url{https://huggingface.co/kevinpro/MistralMathOctopus-7B}} and MetaMathOctopus-7B~\footnote{\url{https://huggingface.co/kevinpro/MetaMathOctopus-7B}}. MistralMathOctopus is obtained by fine-tuning MetaMath-Mistral~\citep{yu2023metamath} with MGSM8KInstruct~\citep{chen2023breaking}. MetaMathOctopus is obtained by fine-tuning MetaMath~\citep{yu2023metamath} with MGSM8KInstruct. Considering limited computational resources and reproducibility, we directly utilize the publicly released base models. Our analyses are mainly based on MistralMathOctopus in the main text and we report more results in the appendix.

\paragraph{Datasets} We conduct experiments on two representative mathematical reasoning benchmarks, MGSM~\citep{shi2022language} and MSVAMP~\citep{chen2023breaking}. MGSM is a widely used benchmark for multilingual mathematical reasoning evaluation. MSVAMP is an out-of-domain test set in contrast to MGSM, which evaluates robustness and generalization~\citep{zhu2024question,she-etal-2024-mapo}.

\paragraph{Languages} Following~\citet{she-etal-2024-mapo}, we choose the following 10 different languages for analysis. As a pivot language, English~(en) is used as the alignment target. We also choose Chinese~(zh), Russian~(ru), German~(de), French~(fr), Spanish~(es), Japanese~(ja), Swahili~(sw), Thai~(th) and Bengali~(bn) as 9 representative non-English languages.

\begin{figure*}[tbp]
    \centering 
    \includegraphics[width=0.48\textwidth]{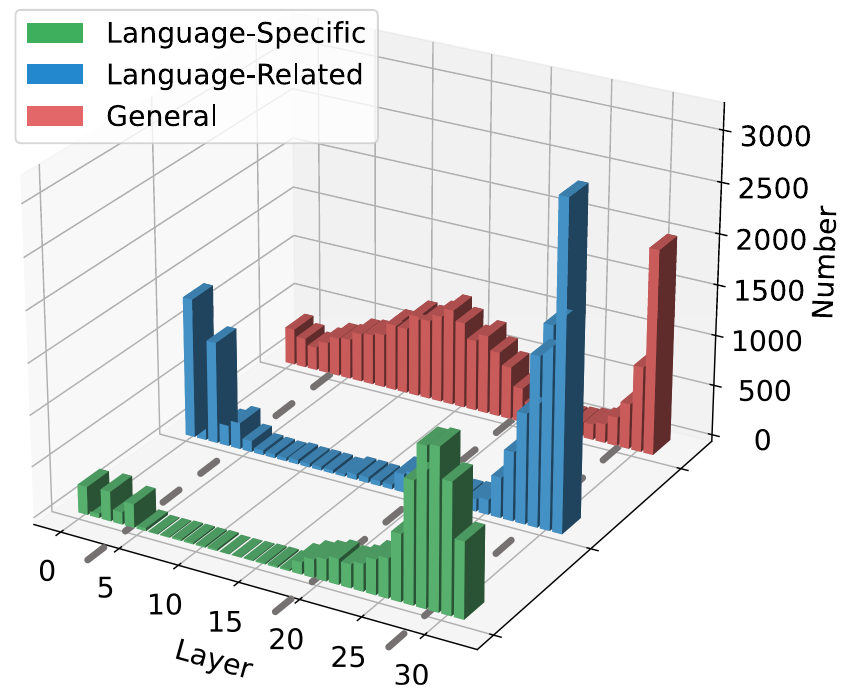}
    \caption{Layer-wise distribution of the different types of neurons of MistralMathOctopus on MGSM.}
    \label{fig:layer-wise-neurons-distribution}
\end{figure*}

\paragraph{Implementations} Due to limited computational resources, our exploration focuses on the most effective DPO variant of MAPO~\citep{she-etal-2024-mapo}. We select 1, 4, and 8 tasks from the NumGLUE~\citep{mishra-etal-2022-numglue}, an arithmetic reasoning benchmark, and translate questions into 9 languages, consistent with the MGSM, thereby creating a multilingual seed dataset.
To construct preference pairs, we sample responses using the corresponding base models and employ NLLB-200-distilled-600M\footnote{\url{https://huggingface.co/facebook/nllb-200-distilled-600M}} as the translation model to obtain alignment scores. Finally, for each model and each target language (excluding English), we gain 10,000 preference pairs. Training is conducted using LoRA~\citep{hu2022lora}.
During the neuron selection stage, we perform force-decoding on the responses of the MGSM or MSVAMP dataset to obtain the activation probabilities of neurons for each language.
Additional implementation details are provided in appendix.

\subsection{Language Neurons Identification}\label{subsec:language-neuron-identification}
Based on the neuron identification algorithm, we identify the language-specific neurons, language-related neurons, and general neurons in the model. To further validate the effectiveness of our algorithm, we report the changes in Accuracy (defined as producing numerically correct answers in the correct language) after deactivating the identified neurons across different languages. Following~\citet{tang-etal-2024-language}, we also report changes in the perplexity~(PPL) scores of LLMs in Appendix. Experiments are conducted on the base model, with results presented in Figure~\ref{fig:mistral-mgsm-heatmap-acc}. We report the results of three settings, deactivating language-specific neurons, deactivating language-specific \& language-related neurons, and deactivating language-specific \& language-related \& general neurons. The higher PPL change (darker cells) indicates the stronger reliance on the neurons be deactivated.

It can be found that \textbf{the performance for each specific language mostly relies on both its language-specific neurons and language-related neurons rather than other neurons}, as the diagonal elements in each row and column show the highest changes in PPL. However, \textbf{different languages vary in the extent to which they rely on their corresponding language-specific and language-related neurons}, reflecting differences in cross-lingual alignment across different languages.

Notably, deactivating both language-specific and language-related neurons leads to a more pronounced effect compared to deactivating only language-specific neurons. Suggesting that for a given language, in addition to its language-specific neurons, \textbf{shared language-related neurons also significantly contribute to its performance}.

Also, deactivating all the language-related neurons of one language doesn't cause significant impacts on the model's performance in other languages, which indicates that the language-related neurons for a specific language are relatively evenly dispersed across multiple other languages. The above findings confirm the validity of the language neurons identified by our method and further provide insights into the characteristics of language neurons.

\subsection{Layer-wise Functionality Analysis}\label{subsec:layer-wise-funtionality-analysis}
Based on the identified neurons, we perform layer-wise functional analyses of all layers in the LLMs. We begin by analyzing the distributions of different types of neurons in the base model. And we report the results in Figure~\ref{fig:layer-wise-neurons-distribution}. 

Some works have divided the LLM's multilingual inference process into three stages~\citep{wendler2024llamas,zhao2024large}. However, through the analysis of the distribution of different types of neurons, we find that although the combined function of the last stage is still to produce language-specific symbols, \textbf{the behavior of language-specific neurons first reaches its peaks and then declines, which differs substantially from the other two types}. 
So we further divide the final stage into two parts, leading to a four-stage interpretation of the LLMs' internal process for multilingual inference:
\begin{enumerate}
    \item \textbf{Multilingual Understanding:} In the initial layers, the number of both language-specific and language-related neurons peaks, while the number of general neurons is relatively low. The model maps multilingual inputs into a unified semantic space at this stage.
    \item \textbf{Shared Semantic Space Reasoning:} In the intermediate layers, the model engages in reasoning within a shared semantic space across different languages. During this stage, both language-specific and language-related neurons are largely absent, whereas general neurons become dominant.
    \item \textbf{Multilingual Output Space Transformation:} The model transfers features into the multilingual output space in this stage in preparation for generating the final output. In this part, the number of both language-specific and language-related neurons reaches a peak again, while the number of general neurons drops to the lowest point.
    \item \textbf{Vocabulary Space Outputting:} In the last layer, the model maps vectors of different languages into a shared vocabulary space to generate outputs. The total number of language-specific neurons and language-related neurons increases, but there is an opposite trend between language-specific and language-related neurons.
    Additionally, unlike assumptions in previous work~\citep{wendler2024llamas,zhao2024large}, the number of general neurons reaches maxima at this stage, consistent with the behavior in the early layers. We therefore speculate that this may be related to the shared vocabulary across different languages.
\end{enumerate}

Meanwhile, the distribution of different types of neurons aligns with the conclusions from the existing studies mentioned above. Overall, the number of neurons varies correspondingly with the different inference stages of LLMs.

\subsection{Layer-wise Neuron Changes Analysis}

\begin{figure}[tbp]
    \centering 
    \includegraphics[width=0.75\columnwidth]{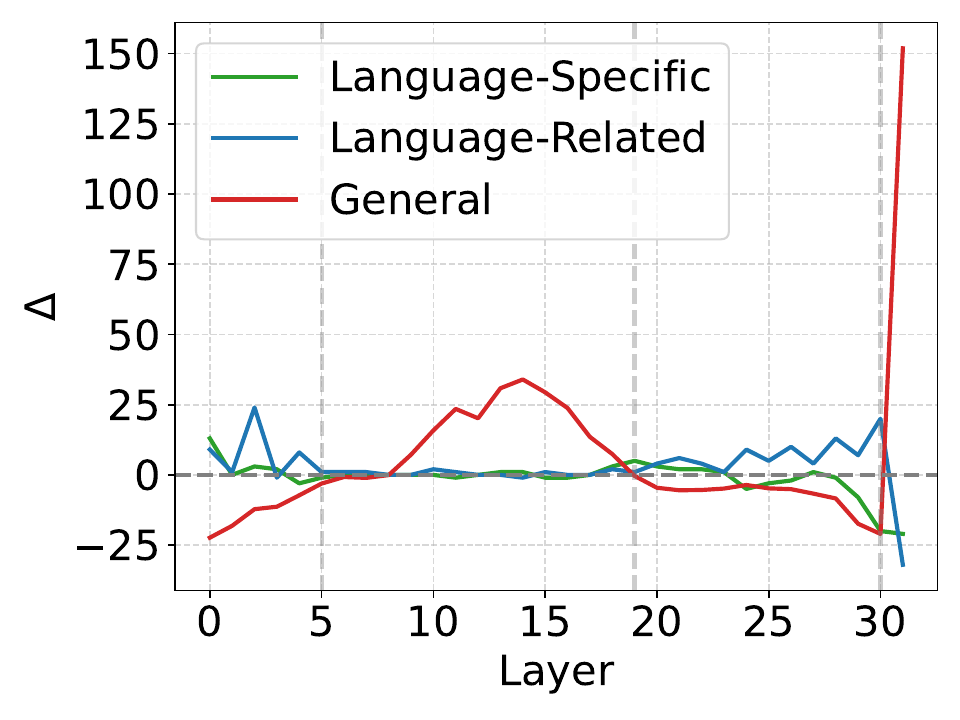}
    \caption{Layer-wise changes in the number of different types of neurons of MistralMathOctopus on MGSM.}
    \label{fig:Delta_of_Language_Neurons}
\end{figure}

We further analyze the changes in different types of neurons before and after alignment. Based on the four functional stages above, we quantify the layer-wise changes~($\Delta$) in the number of different types of neurons in~Figure~\ref{fig:Delta_of_Language_Neurons}.

During the multilingual understanding stage, the number of both language-specific and language-related neurons increases, while general neurons decrease. In the subsequent shared semantic space reasoning stage, general neurons increase substantially, whereas both language-specific neurons and language-related neurons remain stable and nearly absent. 

In the third stage, as general neurons decrease, the sum of language-specific neurons and language-related neurons increase overall. Additionally, we notice that language-specific and language-related neurons show \textbf{opposite trends}. This reflects that the aligned model \textbf{relies more on shared language-related neurons than language-specific neurons in the third stage}, which is difficult to observe under the previous taxonomies. 

Finally, in the last stage, the number of general neurons increases significantly in the aligned model, accompanied by a reduction in both language-specific and language-related neurons. This \textbf{contradicts the conventional three-stage assumption}, indicating that general neurons also play a significant role in the final vocabulary outputting stage. Additionally, combined with the observations in Figure~\ref{fig:Related_Language_Num}, the model relies on a smaller set of language neurons with a higher degree of sharing in this stage. We also report the results of different checkpoints during the alignment process in appendix.

Overall, we find the \textit{union set} of language-specific and language-related neurons and \textit{general neurons} exhibit generally opposite trends across layers, which corresponds to the characteristics of LLMs at different stages. Especially, at the last stage, general neurons play an more important role. Multilingual alignment facilitates more effective activation of the appropriate neurons at each stage, thereby improving the model's capability to handle multilingual tasks.

\subsection{Macroscopic Analysis of Neurons}
We further conduct macroscopic analysis for different types of neurons. In our neuron identification algorithm, the number of languages that share a specific neuron is an attribution characterizing all types of activated neurons. Since our study involves 10 languages, the valid range of $N$ is from $1$ to $10$. Among these, values of $N$ from $2$ to $9$ correspond to language-related neurons. As special cases in our work, $N = 1$ represents language-specific neurons, while $N = 10$ corresponds to general neurons.

\begin{figure}[tbp]
    \centering 
    \includegraphics[width=0.93\columnwidth]{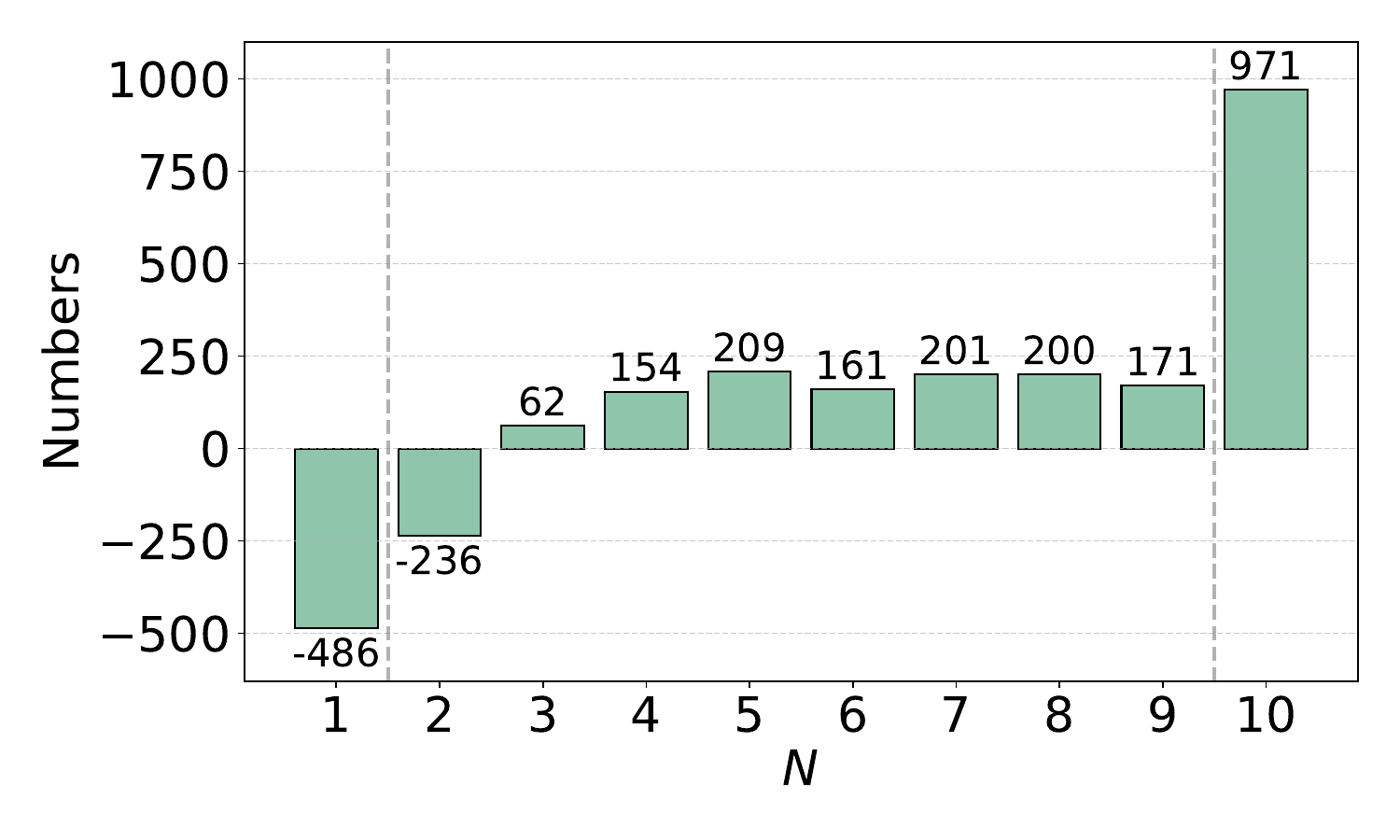}
    \caption{Changes in the number of neurons shared by $N$ languages after alignment.}
    \label{fig:Related_Language_Num}
\end{figure}

We report the changes in the number of neurons after multilingual alignment for each value of $N$ ranging from 1 to 10 in Figure~\ref{fig:Related_Language_Num}. The results show a decrease in the number of language-specific neurons, while an increase in the number of language-related neurons,  which are shared across multiple languages. This indicates that \textbf{multilingual alignment encourages LLMs to develop and utilize more shared language-related neurons, rather than language-specific neurons}, which are applicable to only a single language. Meanwhile, during the alignment, the model improves its task-relevant common knowledge. Therefore, the number of general neurons also increases significantly. Language-related neurons and language-specific neurons exhibit overall opposite trends, further highlighting the importance of a more fine-grained analysis of the two categories within our new taxonomy. In addition, we also report the overall neuron numbers for each value of $N$ before and after alignment in Appendix.

\begin{table*}[tbp]
\centering
\small
\resizebox{0.7\textwidth}{!}{
\begin{tabular}{l|cccccccccc|c}
\toprule
\textbf{MGSM} & \textbf{bn} & \textbf{th} & \textbf{sw} & \textbf{ja} & \textbf{zh} & \textbf{ru} & \textbf{de} & \textbf{es} & \textbf{fr} & \textbf{en} & \textbf{Avg.} \\
\midrule
base & 43.6 & 53.2 & 50.4 & 55.6 & 59.6 & 59.2 & 61.2 & 62.8 & 56.8 & 75.6 & 57.8 \\
zh/de $\Rightarrow$ en & 46.4 & 55.6 & 59.2 & 56.8 & 64.0 & 71.2 & 66.8 & 71.2 & 69.2 & 75.2 & 63.6 \\
sw/th $\Rightarrow$ en & 48.8 & 58.8 & 59.2 & 56.4 & 68.4 & 68.4 & 69.2 & 69.6 & 70.4 & 77.6 & 64.7 \\
\bottomrule
\end{tabular}
}
\caption{Accuracy of the MistralMathOctopus base model and aligned model on MGSM. "X/Y $\Rightarrow$ T" indicates that languages X and Y are used for multilingual alignment.}
\label{tab:spontaneous_result}
\end{table*}

\subsection{Spontaneous Multilingual Alignment Analysis}\label{subsec:spontaneous-multilingual-alignment-analysis}
The ``spontaneous multilingual alignment'' phenomenon is first revealed and discussed by~\citet{zhang-etal-2024-getting}, which shows that conducting alignment in a small number of languages significantly improves multilingual alignment even between English and many languages unseen during the alignment process. We further analyze this phenomenon in our experiments. As shown in Table~\ref{tab:spontaneous_result}, spontaneous multilingual alignment also emerges under the multilingual alignment method employed in our study. Besides the languages used for alignment, LLMs exhibit notable performance gains in other unaligned languages.
To understand how multilingual alignment generalizes to other languages, we analyze the changes in different types of neurons before and after multilingual alignment based on our method.

\begin{table}[tbp]
\centering
\small
\resizebox{\columnwidth}{!}{
\begin{tabular}{lcc}
\toprule
\textbf{Language} & \textbf{Language-Specific} & \textbf{Language-Related} \\
\midrule
Trained & -37 & +232 \\
Others & -36 & +205 \\
\bottomrule
\end{tabular}
}
\caption{Average results of neuron count changes across multiple languages. ``Trained'' indicates the trained languages in the spontaneous multilingual alignment experiment. ``Others'' indicates other languages except the trained languages. We round the results to the nearest integer.}
\label{tab:spontaneous_delta_neurons_zhde}
\end{table}

Taking the case of ``zh/de $\Rightarrow$ en'' as a representative example, we report the average results in Table~\ref{tab:spontaneous_delta_neurons_zhde}. For the trained languages, the number of language-specific neurons decreases, while the number of language-related neurons increases. This indicates the aligned languages tend to utilize more language-related neurons shared with other languages rather than exclusive language-specific neurons. Moreover, we extend this analysis to languages other than the trained languages and observe a similar phenomenon. These findings indicate \textbf{multilingual alignment facilitates the use of language-related neurons while reducing the reliance on language-specific neurons in both trained and other unseen languages}. We hypothesize that the new language-related neurons shared with trained languages contribute to the performance improvement on other unseen languages.

\begin{table}[tbp]
    \centering
    \footnotesize
    \resizebox{\columnwidth}{!}{\begin{tabular}{lcc}
    \toprule
    \textbf{Language} & \textbf{Language-Specific} & \textbf{Language-Related} \\
    \midrule
    English & 46 & 603 \\
    non-English & 613 & 2006 \\
    \bottomrule
    \end{tabular}}
    \caption{{Average number of different types of neurons for English and non-English languages of MistralMathOctopus on MGSM. We round the results to the nearest integer.}}
    \label{tab:English-nonEnglish-neuron-count}
\end{table}

\subsection{Further Analysis}\label{subsec:further-analysis}

\paragraph{Uniqueness of English}
Since current LLMs are primarily pretrained on English data, English is often regarded as playing a special role within LLMs~\citep{wendler2024llamas}. In our work, we observe that English exhibits markedly different characteristics compared to other languages. Based on the identified neurons in our work, in Figure~\ref{fig:mistral-mgsm-heatmap-acc}, deactivating the language neurons of English has a negligible impact on the model's performance in English, which is entirely different from the behavior observed in other languages. This is consistent with the results of~\citet{tang-etal-2024-language}. Furthermore, we quantify the sum of language-specific neurons and language-related neurons for English and non-English languages based on the MistralMathOctopus base model~(Table~\ref{tab:English-nonEnglish-neuron-count}). 

Our analysis reveals that English has significantly fewer neurons than other languages, both in terms of language-specific and language-related neurons. We hypothesize that this is due to that English actually possesses numerous language-related neurons. And since English serves as a pivot language, these language-related neurons are likely shared with almost all other languages. It causes them to be also regarded as general neurons.


\begin{table}[tbp]
    \centering
    \footnotesize
    \resizebox{\columnwidth}{!}{
    \begin{tabular}{lcc}
    \toprule
    \textbf{Variable~(\%)} & \textbf{Language-Specific} & \textbf{Language-Related} \\
    \midrule
    Domain & 80.7 & 92.3 \\
    Alignment & 95.6 & 92.1 \\
    \bottomrule
    \end{tabular}
    }
    \caption{Overlap ratio of different types of neurons across \textit{different domains} and \textit{before and after alignment}. Following~\citet{she-etal-2024-mapo}, MSVAMP is regarded as an out-of-domain dataset. The results of MistralMathOctopus on MGSM are used as the fiducial value.}
    \label{tab:neuron-distribution-stability-overlap}
\end{table}

\paragraph{Stability of Neuron Distributions}We discuss the stability of neuron distributions across \textit{different data domains}, as well as \textit{before and after alignment}. To quantify the stability of neuron distributions, we compute the neuron overlap ratio in both settings, with the results summarized in Table~\ref{tab:neuron-distribution-stability-overlap}. 
We can find that although the exact positions of a few neurons may vary across different settings, the positional distribution of most neurons remains stable. This also indicates good reliability and generalization of the language neurons identified under fixed hyperparameters.


\section{Conclusion}\label{sec:conclusion}

In this work, we systematically investigate the multilingual alignment from the perspective of language neurons. First, based on the defects of the existing binary classification methodology, we propose a ternary classification methodology, which defines the language-specific neurons, language-related neurons, and general neurons. Then we propose a corresponding language neuron identification algorithm, which detects the above different types of neurons in LLMs. 

Furthermore, we examine the multilingual alignment mechanism by analyzing the roles of different types of neurons. 
Based on their distributional characteristics, we categorize LLMs' internal process into four functional parts. Our analysis reveals that multilingual alignment enhances the model’s utilization of the corresponding types of neurons across different functional parts. Meanwhile, we find that alignment promotes a greater reliance on shared language-related neurons across languages, rather than on language-specific neurons. This also corresponds to the phenomenon of spontaneous multilingual alignment. 

Overall, we provide further analysis and valuable insights to better understand multilingual alignment and multilingual capabilities of LLMs.

\section{Acknowledgements}\label{sec:ack}
We would like to thank the anonymous reviewers for their insightful comments. Shujian Huang is the corresponding author. This work is supported by National Science Foundation of China (No. 62376116), research project of Nanjing University-China Mobile Joint Institute (NJ20250038), the Fundamental Research Funds for the Central Universities (No. 2024300507, 2025300390).


\bibliography{aaai2026}


\newpage
\appendix

\section{Implementation Details}\label{app:implementation-details}
Our experiments are conducted on 4 NVIDIA RTX A6000 GPUs or 4 NVIDIA GeForce RTX 3090 GPUs. We use the TRL~\citep{vonwerra2022trl} and DeepSpeed~\citep{rasley2020deepspeed} frameworks for preference alignment, and the vLLM engine~\cite{kwon2023efficient} for inference.

\subsection{MAPO}
We employ the officially released scripts\footnote{\url{https://github.com/NJUNLP/MAPO}} to generate the preference data. For the alignment process, we set the learning rate to 1e-6 and the batch size to 16. LoRA is utilized to fine-tune the model with a LoRA rank of 64, a LoRA alpha of 128, and a LoRA dropout rate of 0.05. The total number of training steps is set to 1000. It takes 7.5 hours for one alignment.

\subsection{Language Neurons Identification}
We spend 25 minutes calculating the activation probability of each neuron with MGSM dataset on a single NVIDIA GeForce RTX 3090 GPU. 
We spend 2 hours calculating the activation probability of each neuron with MSVAMP dataset on a single NVIDIA GeForce RTX 3090 GPU.
After obtaining the activation probability, it takes 3 minutes to identify how many languages each selected neuron is related to.

\section{Licenses for Used Assets}\label{app:licenses}
We list the names of the licenses for each asset we utilized in our work:
\begin{itemize}
    \item MGSM: CC-BY-SA-4.0
    \item MSVAMP: Apache-2.0
    \item GSM8KInstruct: Apache-2.0
    \item NumGLUE: Apache-2.0
    \item MistralMathOctopus: Apache-2.0
    \item MetaMathOctopus: Apache-2.0
    \item NLLB-200-distilled-600M: CC-BY-NC-4.0
    \item TRL: Apache-2.0
    \item DeepSpeed: Apache-2.0
    \item vLLM: Apache-2.0
\end{itemize}

\section{Deactivation Ablation Experiments}
\label{app:heatmap}
\begin{figure*}[htbp]
    \centering 
    \begin{minipage}[b]{0.41\textwidth}
    \includegraphics[width=1.0\linewidth]{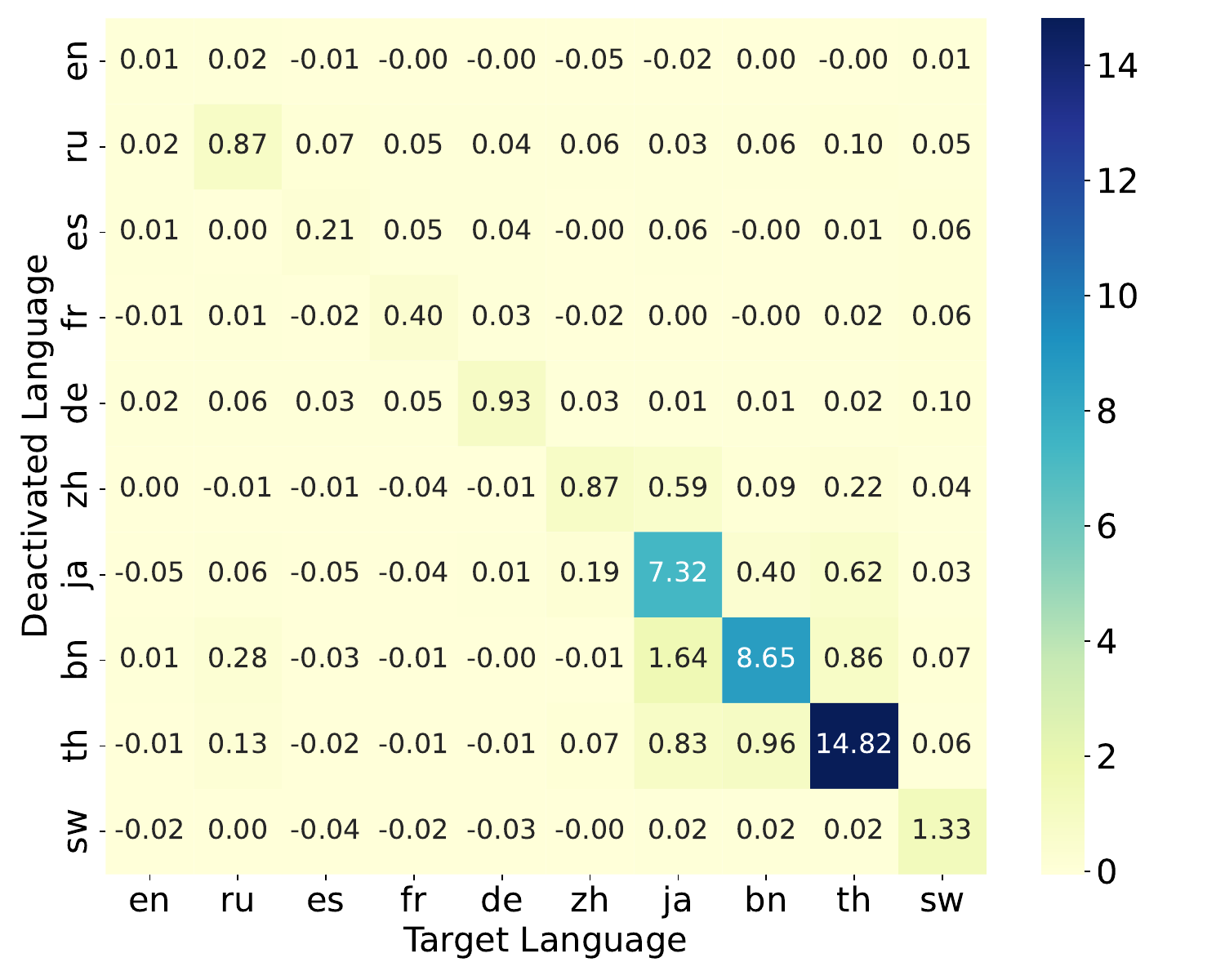}
    \subcaption{Base - Language-specific Neurons}
    \label{fig:mistral-mgsm-baseline-base-heatmap}
    \end{minipage}
    \begin{minipage}[b]{0.41\textwidth}
    \includegraphics[width=1.0\linewidth]{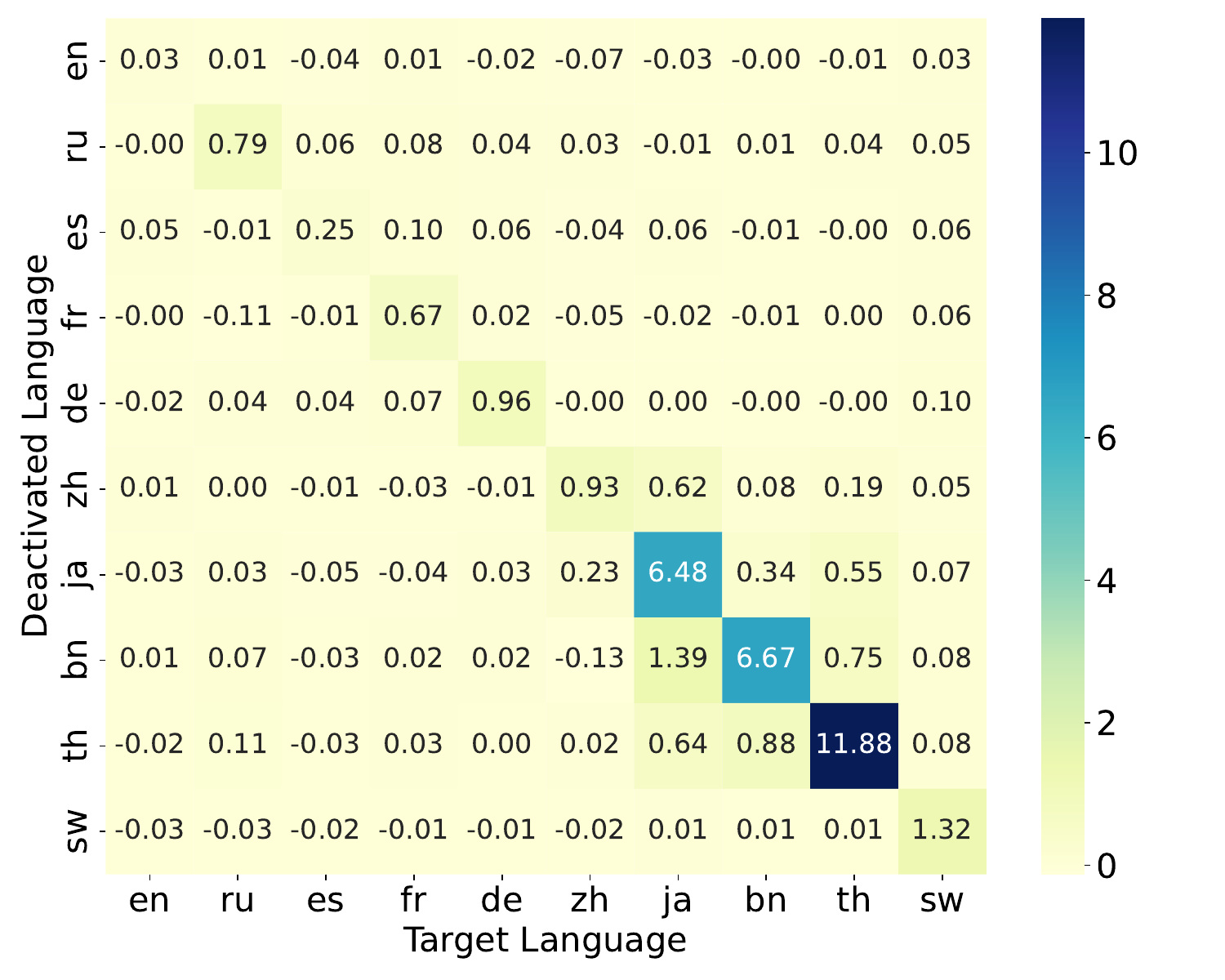}
    \subcaption{Aligned - Language-specific Neurons}
    \label{fig:mistral-mgsm-baseline-aligned-heatmap}
    \end{minipage}
    \caption{PPL changes of MistralMathOctopus on MGSM when deploying the algorithm proposed by~\citet{tang-etal-2024-language}.}
    \label{fig:baseline-heatmap}\
    \vspace{-10pt}
\end{figure*}

\begin{figure*}[htbp]
    \centering
    \begin{minipage}[b]{0.33\textwidth}
    \includegraphics[width=1.0\linewidth]{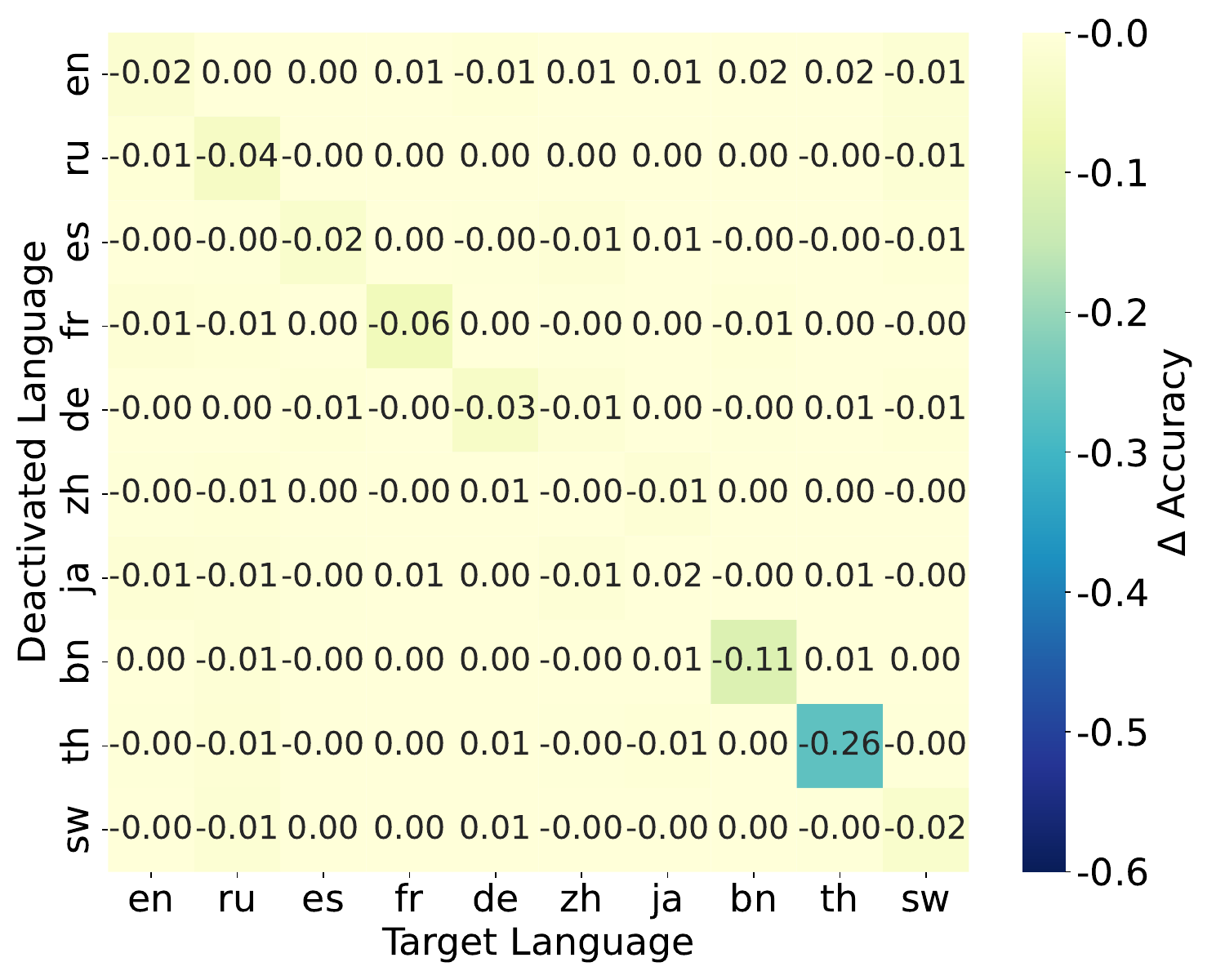}
    \subcaption{Specific Neurons}
    \end{minipage}
    \begin{minipage}[b]{0.33\textwidth}
    \includegraphics[width=1.0\linewidth]{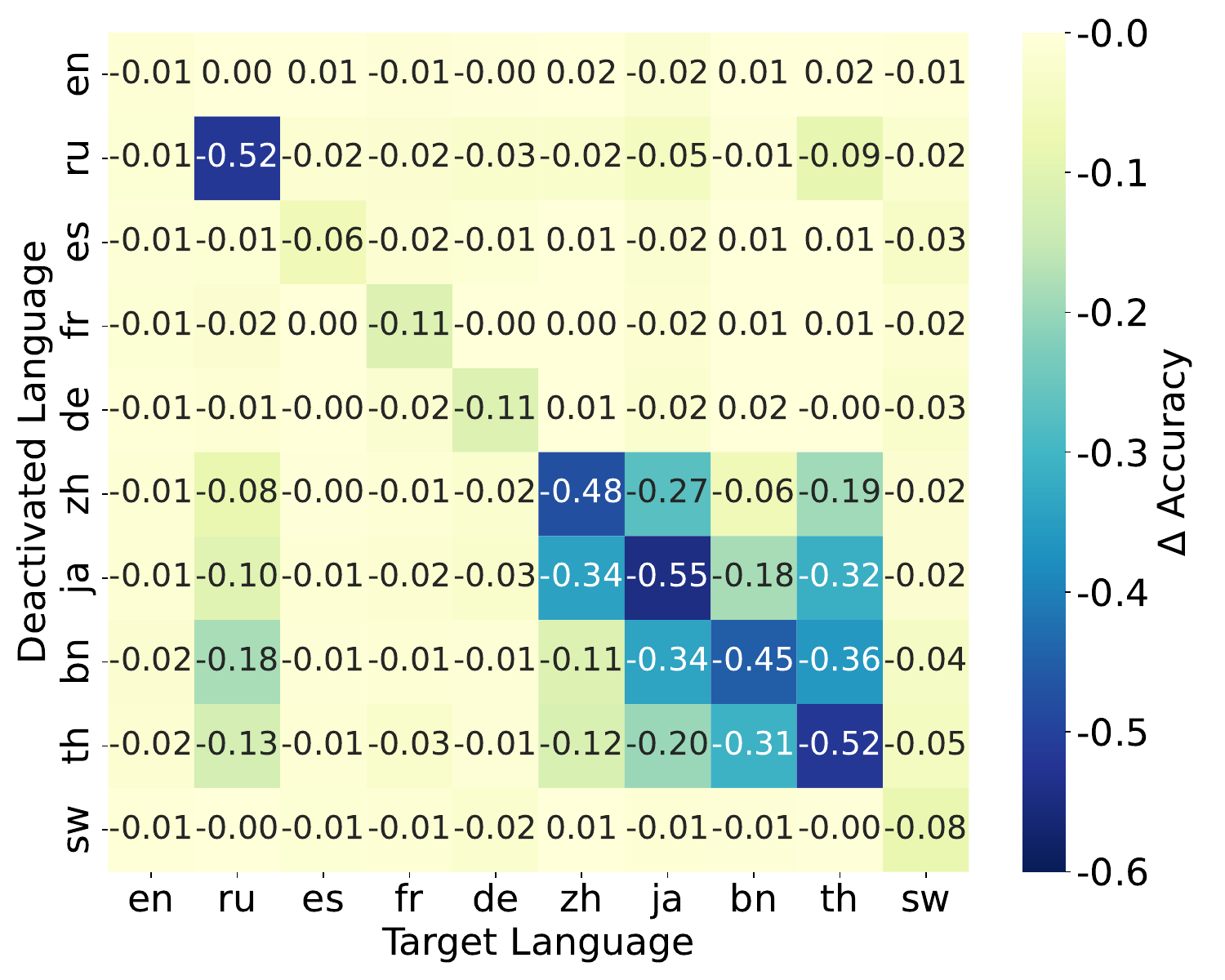}
    \subcaption{Specific \& Related Neurons}
    \end{minipage}
    \begin{minipage}[b]{0.33\textwidth}
    \includegraphics[width=1.0\linewidth]{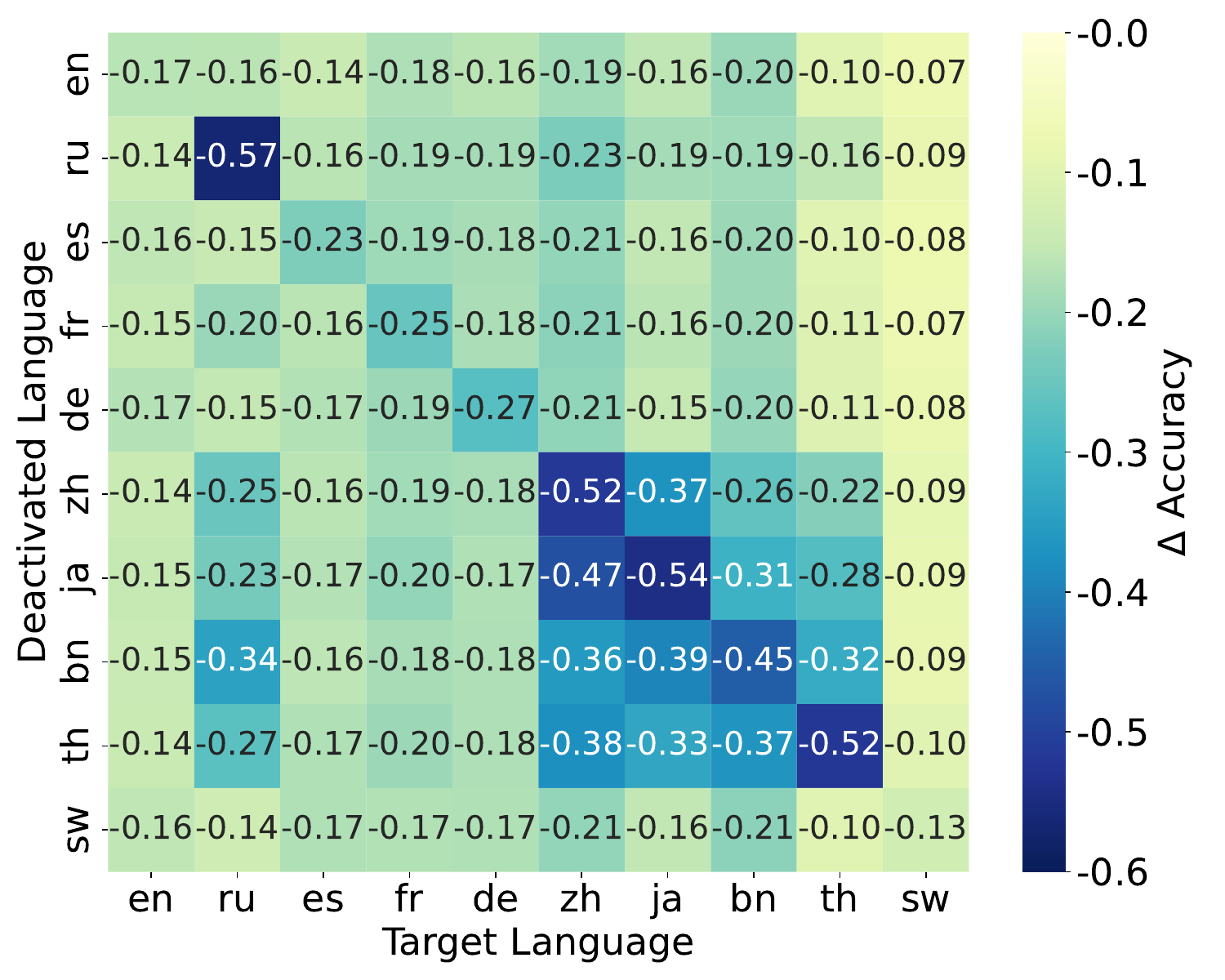}
    \subcaption{Specific \& Related \& General Neurons}
    \end{minipage}
\caption{Accuracy changes of MistralMathOctopus on MSVAMP after deactivating language-specific neurons or language-specific \& language-related neurons or language-specific \& language-related \& general neurons.}
\label{fig:mistral-msvamp-heatmap-acc}
\end{figure*}

\begin{figure*}[htbp]
    \centering
    \begin{minipage}[b]{0.33\textwidth}
    \includegraphics[width=1.0\linewidth]{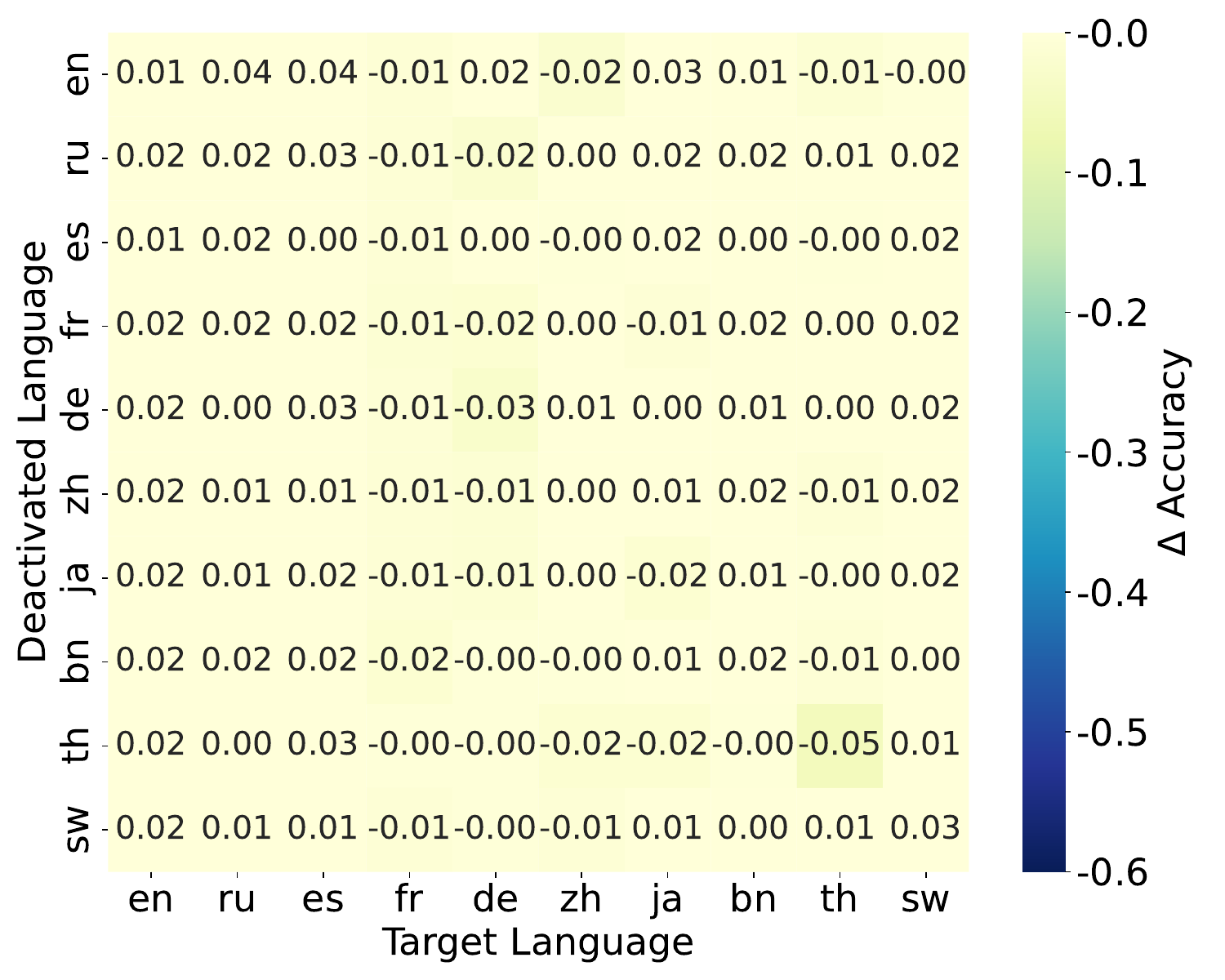}
    \subcaption{Specific Neurons}
    \end{minipage}
    \begin{minipage}[b]{0.33\textwidth}
    \includegraphics[width=1.0\linewidth]{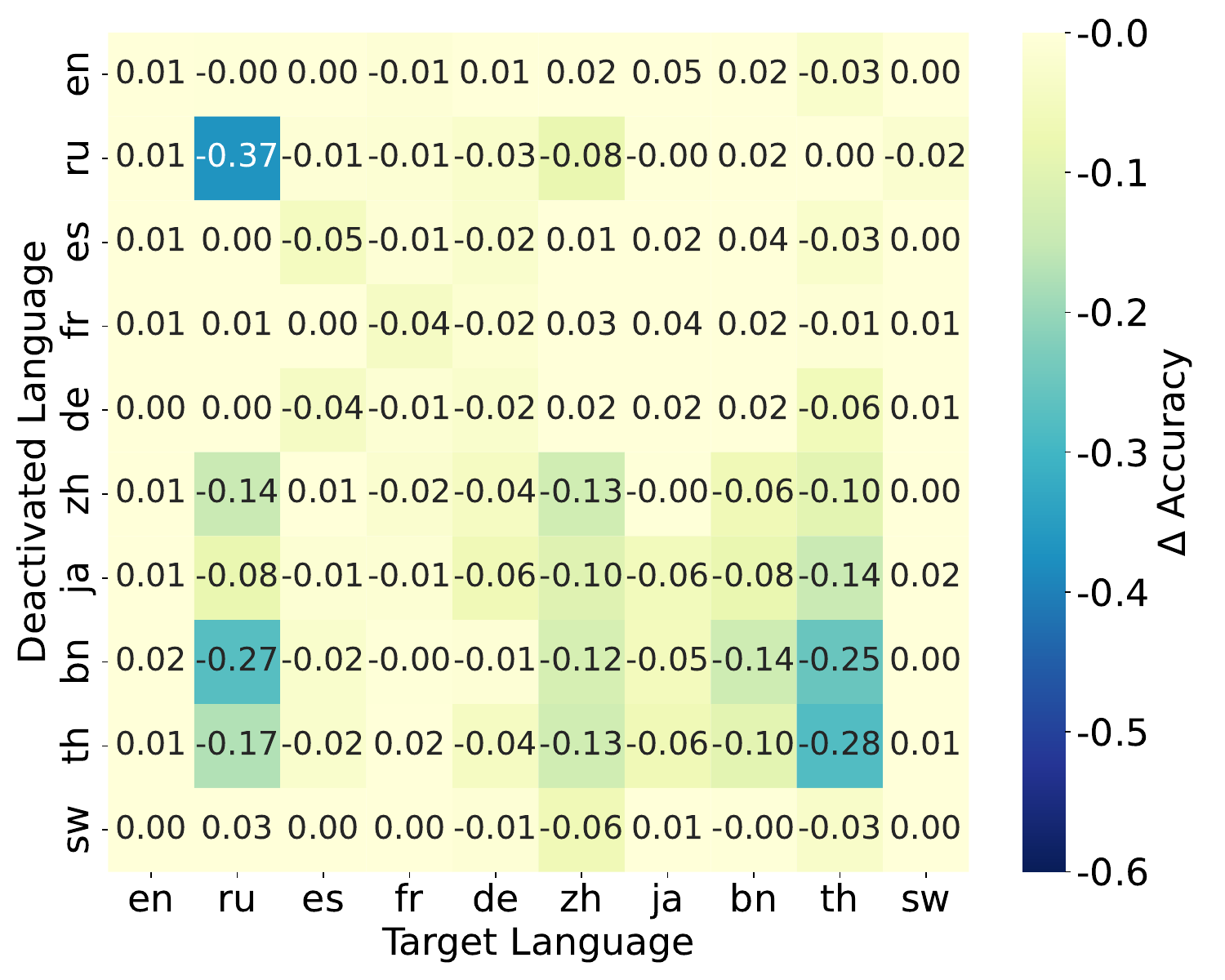}
    \subcaption{Specific \& Related Neurons}
    \end{minipage}
    \begin{minipage}[b]{0.33\textwidth}
    \includegraphics[width=1.0\linewidth]{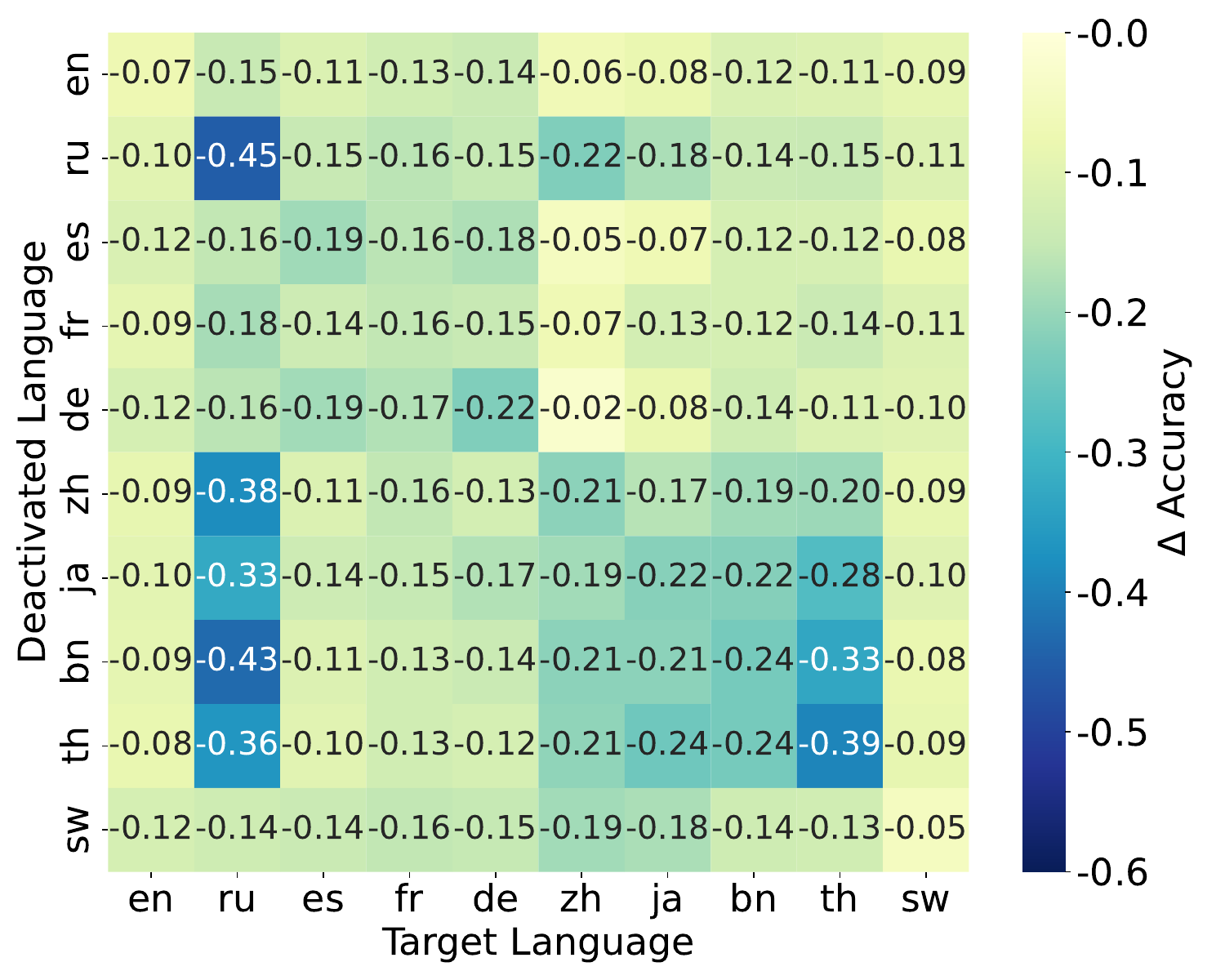}
    \subcaption{Specific \& Related \& General Neurons}
    \end{minipage}
\caption{Accuracy changes of MetaMathOctopus on MGSM after deactivating language-specific neurons or language-specific \& language-related neurons or language-specific \& language-related \& general neurons.}
\label{fig:llama-mgsm-heatmap-acc}
\end{figure*}

\begin{figure*}[htbp]
    \centering
    \begin{minipage}[b]{0.33\textwidth}
    \includegraphics[width=1.0\linewidth]{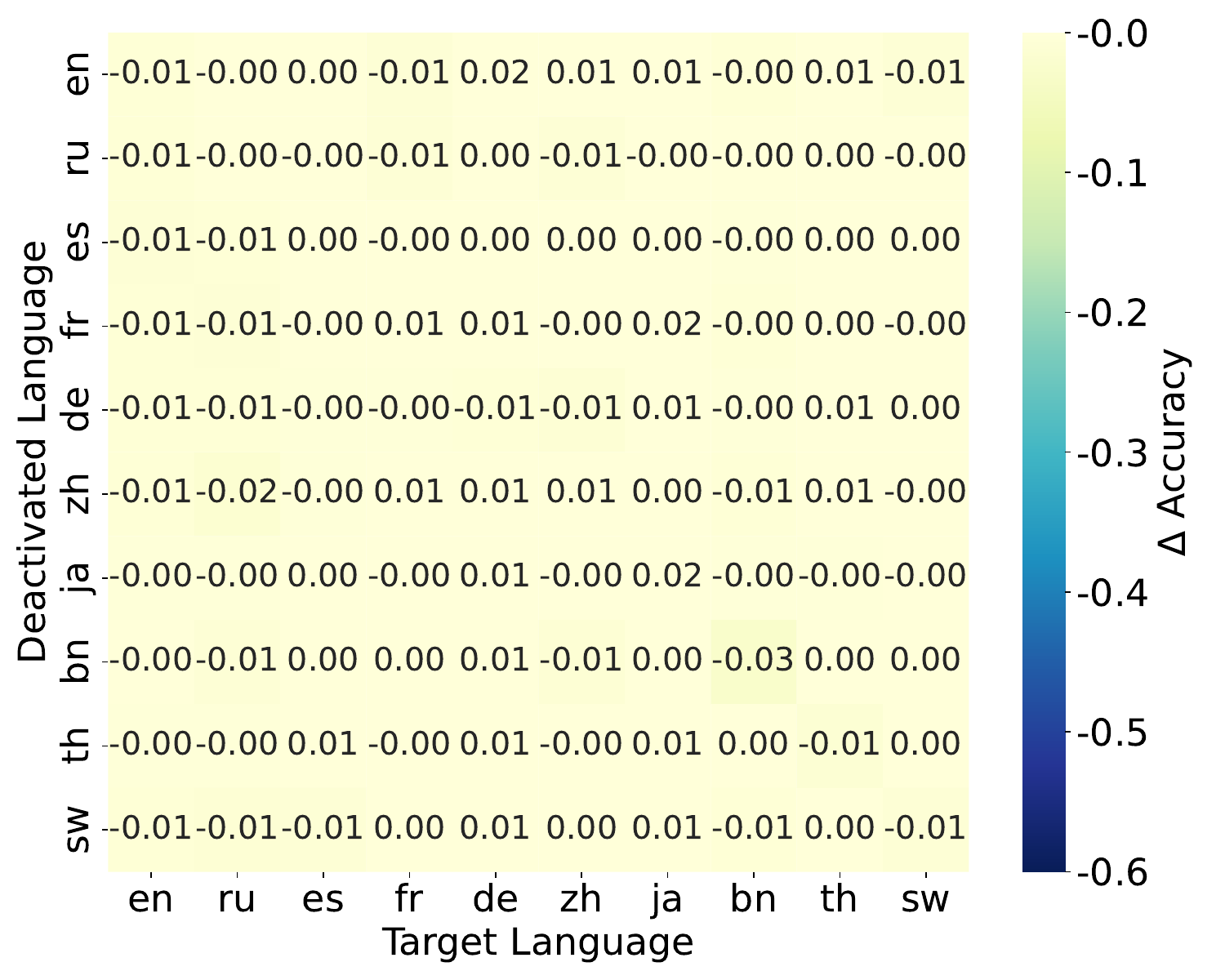}
    \subcaption{Specific Neurons}
    \end{minipage}
    \begin{minipage}[b]{0.33\textwidth}
    \includegraphics[width=1.0\linewidth]{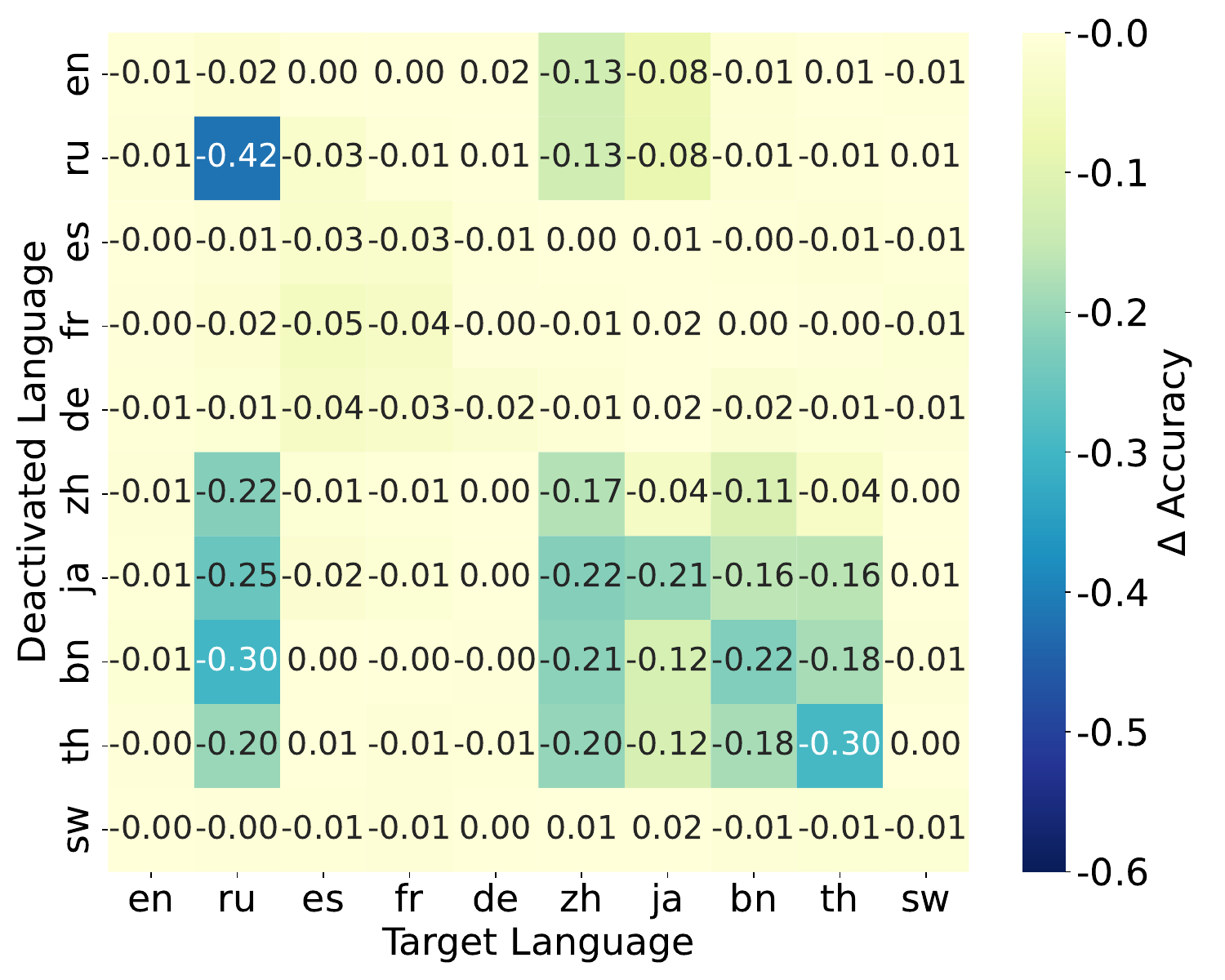}
    \subcaption{Specific \& Related Neurons}
    \end{minipage}
    \begin{minipage}[b]{0.33\textwidth}
    \includegraphics[width=1.0\linewidth]{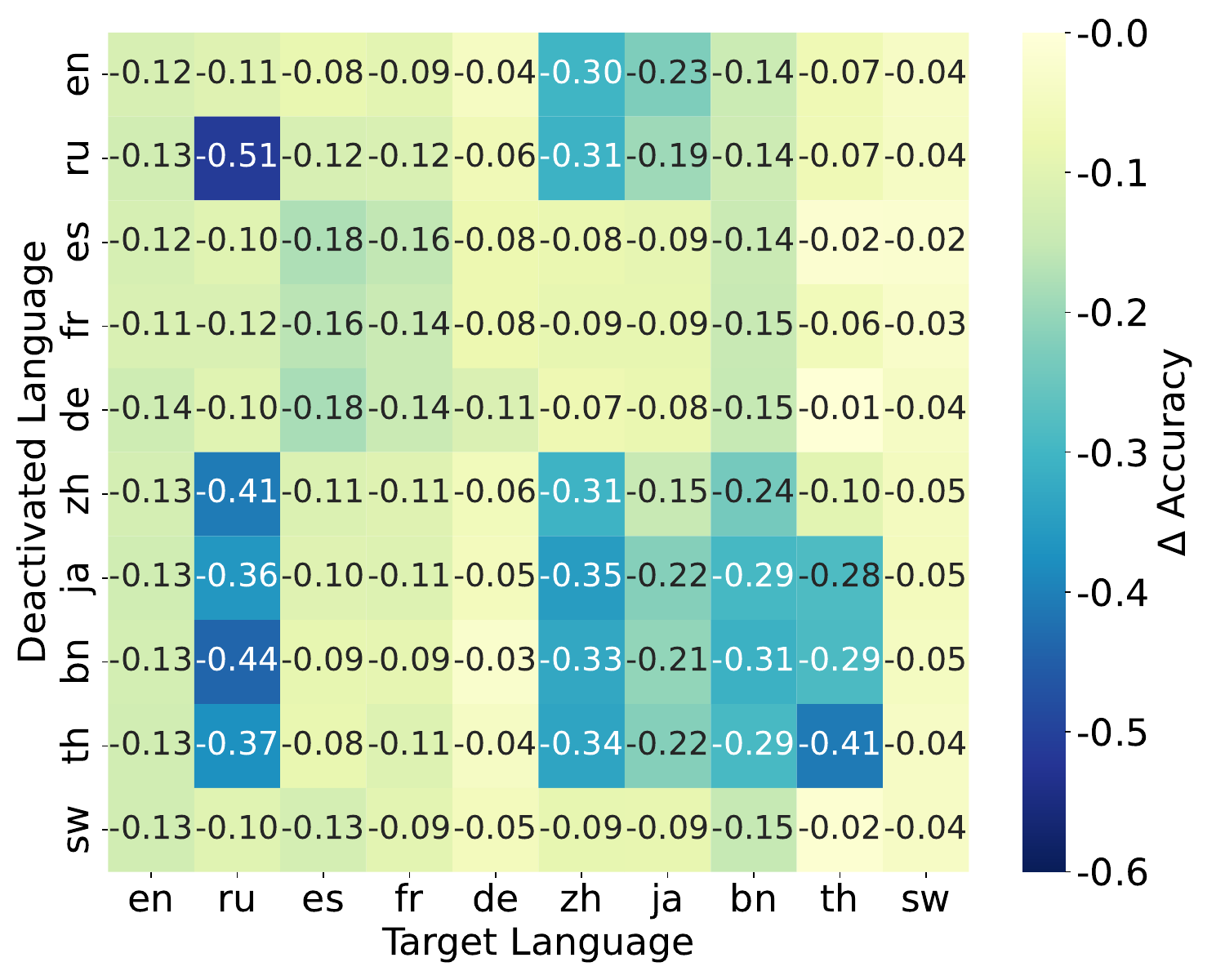}
    \subcaption{Specific \& Related \& General Neurons}
    \end{minipage}
\caption{Accuracy changes of MetaMathOctopus on MSVAMP after deactivating language-specific neurons or language-specific \& language-related neurons or language-specific \& language-related \& general neurons.}
\label{fig:llama-msvamp-heatmap-acc}
\end{figure*}

\begin{figure*}[htbp]
    \centering
    \begin{minipage}[b]{0.33\textwidth}
    \includegraphics[width=1.0\linewidth]{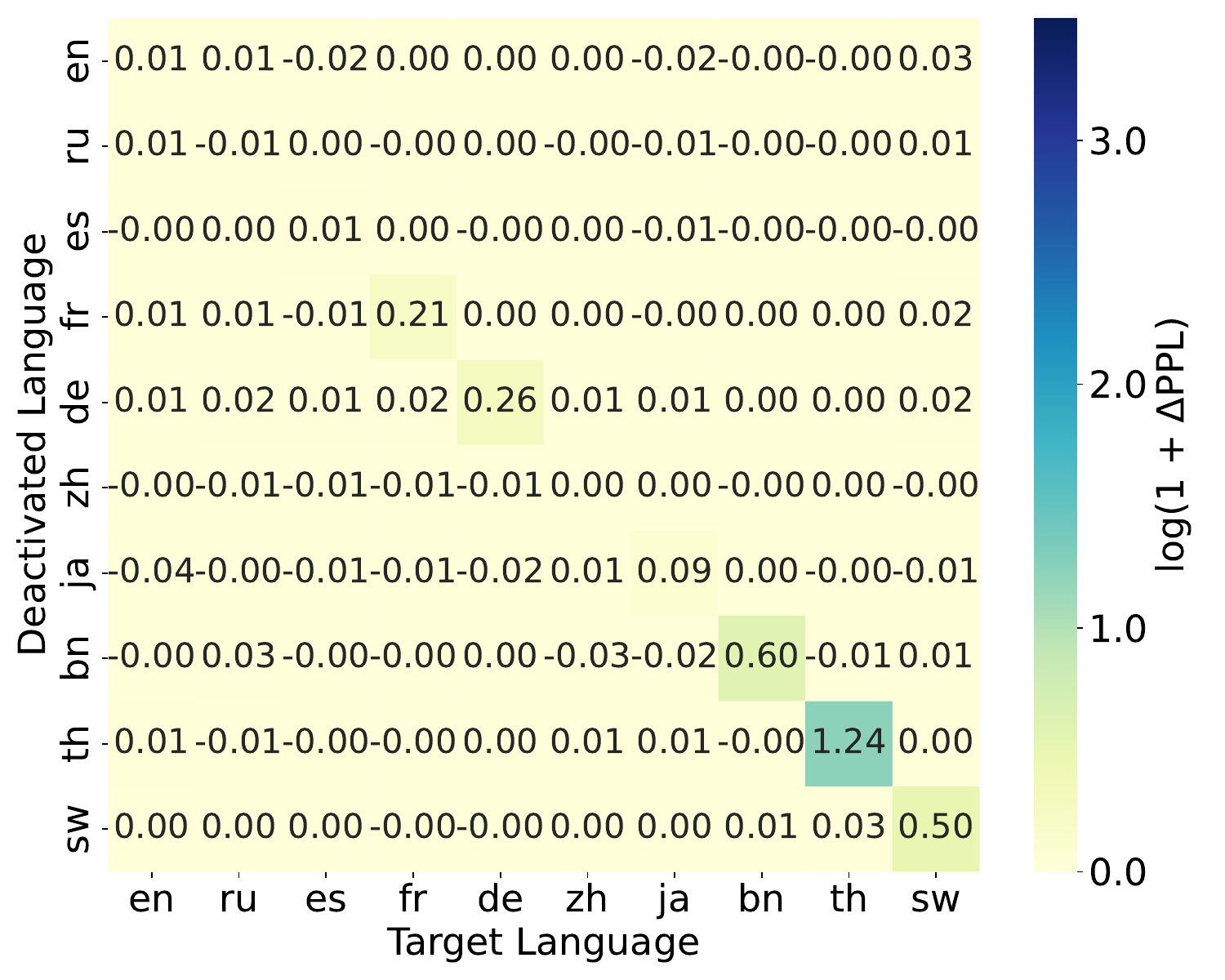}
    \subcaption{Specific Neurons}
    \end{minipage}
    \begin{minipage}[b]{0.33\textwidth}
    \includegraphics[width=1.0\linewidth]{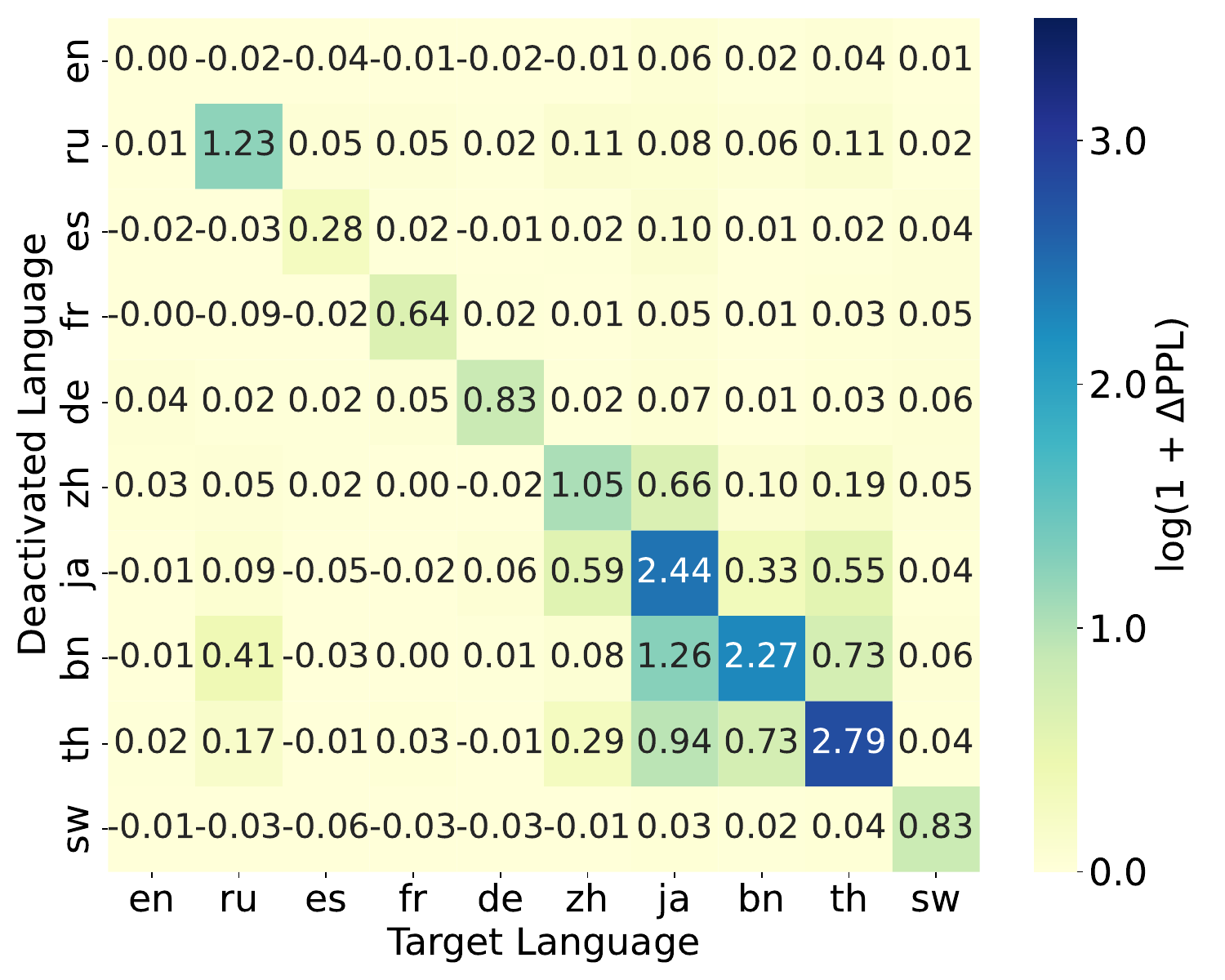}
    \subcaption{Specific \& Related Neurons}
    \end{minipage}
    \begin{minipage}[b]{0.33\textwidth}
    \includegraphics[width=1.0\linewidth]{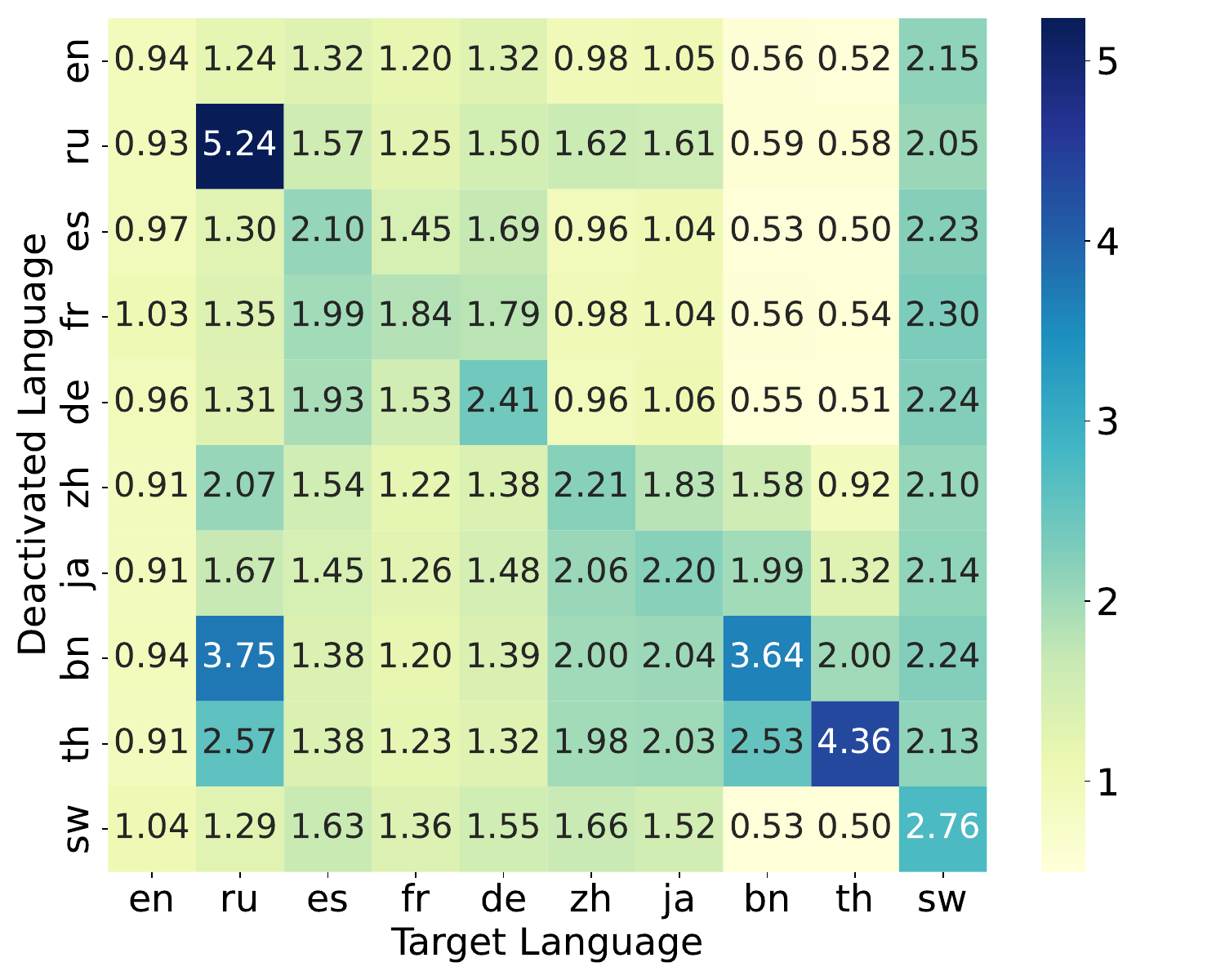}
    \subcaption{Specific \& Related \& General Neurons}
    \end{minipage}
\caption{PPL changes of MistralMathOctopus on MGSM after deactivating language-specific neurons or language-specific \& language-related neurons or language-specific \& language-related \& general neurons.}
\label{fig:mistral-mgsm-heatmap-ppl}
\end{figure*}

\begin{figure*}[htbp]
    \centering
    \begin{minipage}[b]{0.33\textwidth}
    \includegraphics[width=1.0\linewidth]{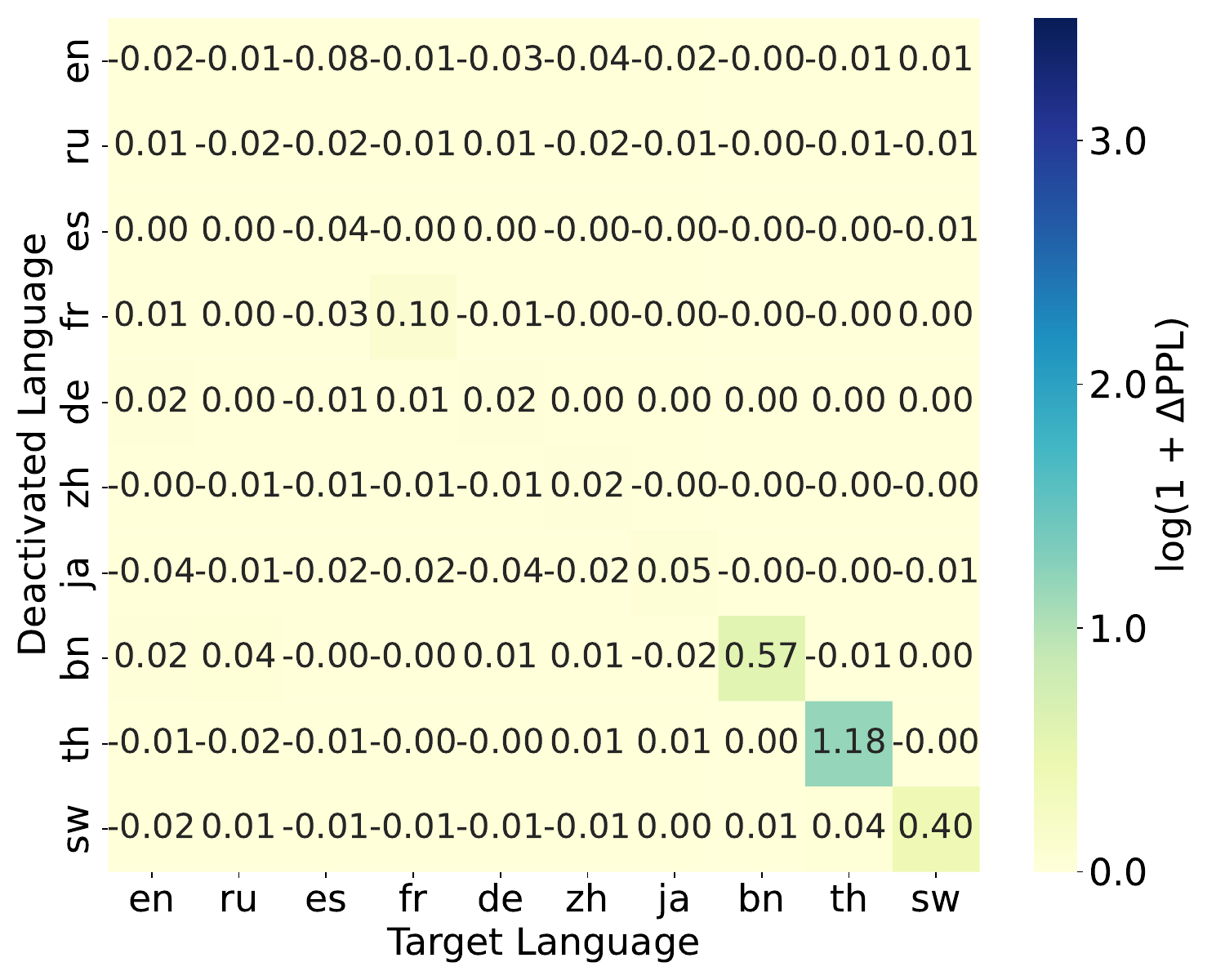}
    \subcaption{Specific Neurons}
    \end{minipage}
    \begin{minipage}[b]{0.33\textwidth}
    \includegraphics[width=1.0\linewidth]{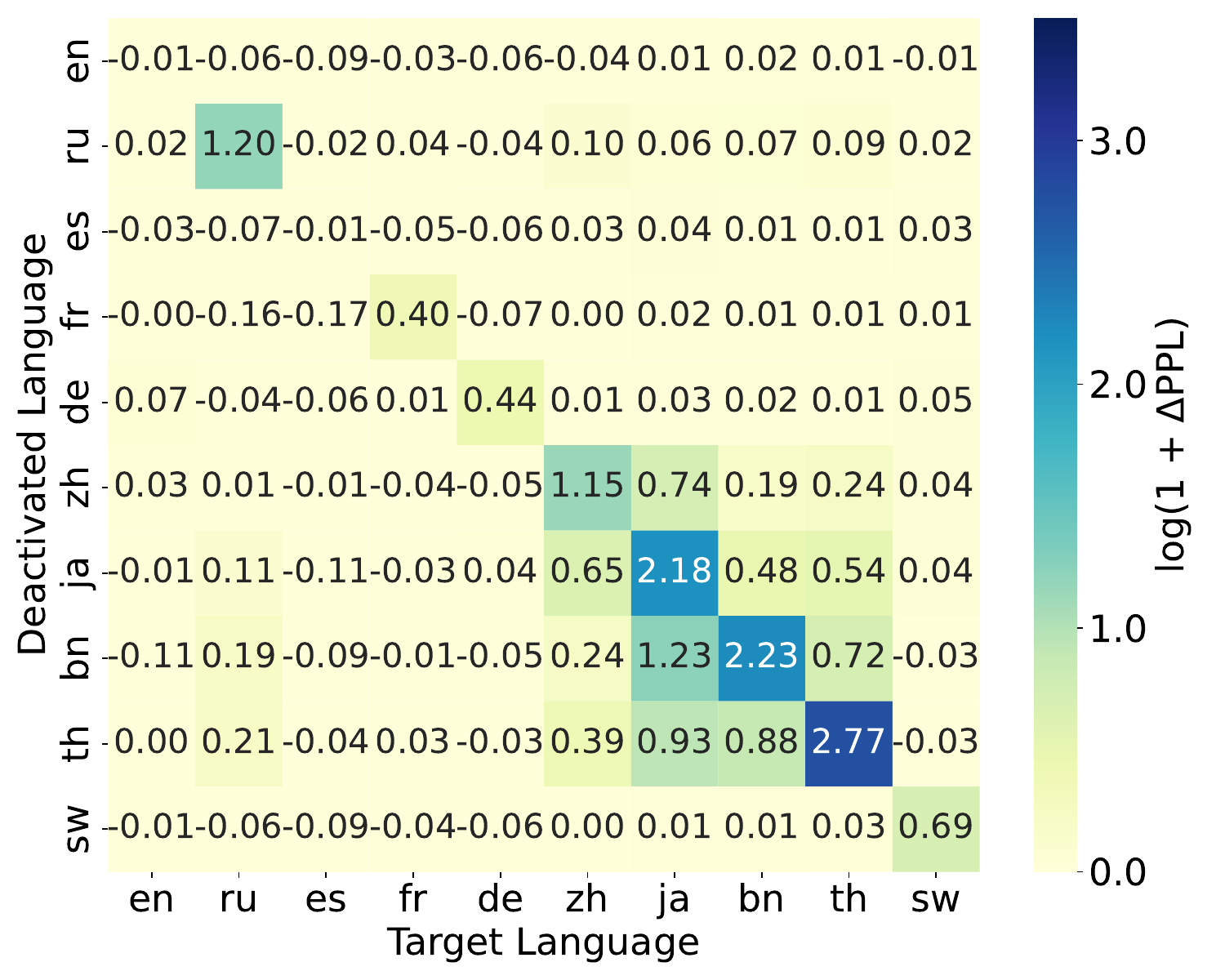}
    \subcaption{Specific \& Related Neurons}
    \end{minipage}
    \begin{minipage}[b]{0.33\textwidth}
    \includegraphics[width=1.0\linewidth]{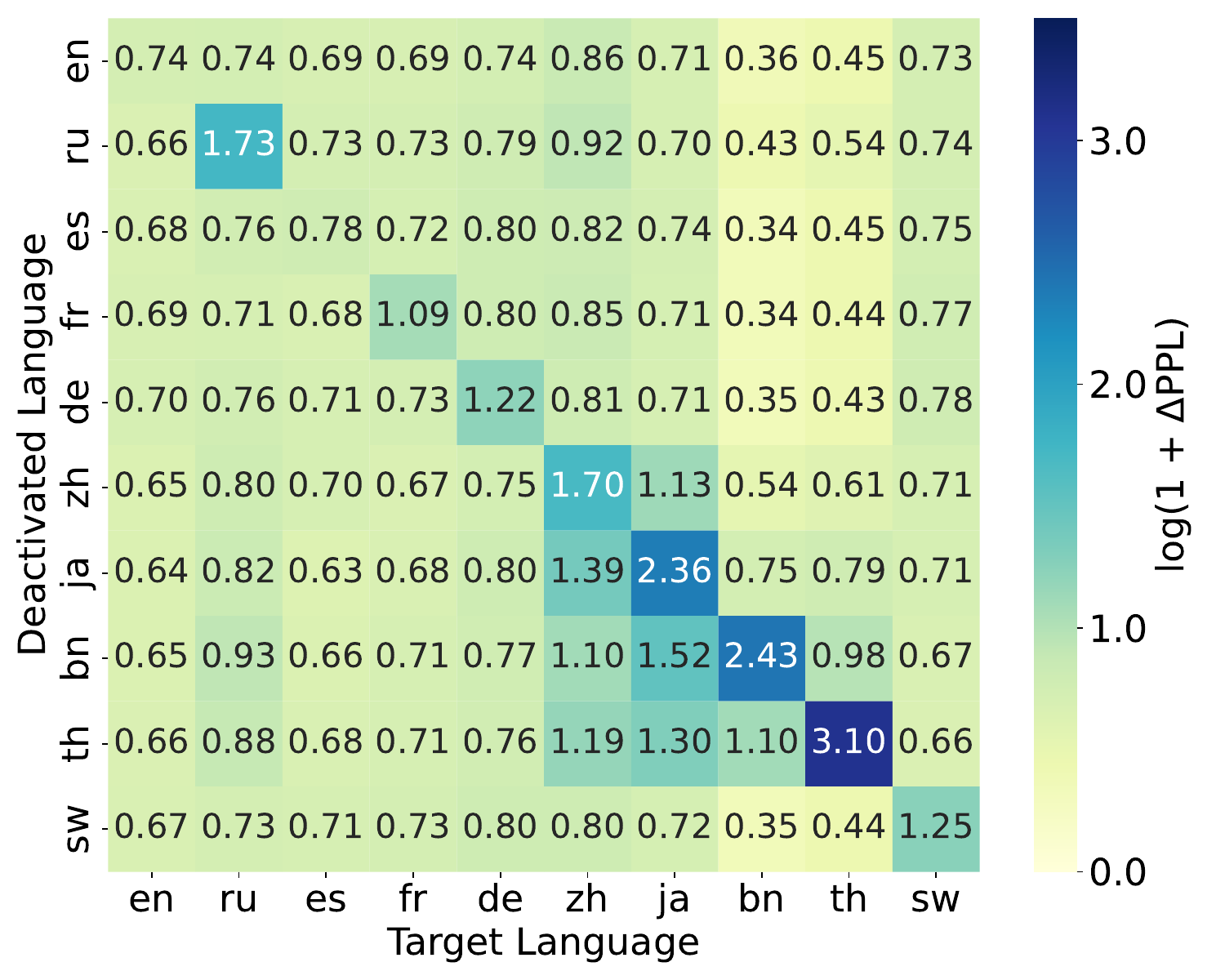}
    \subcaption{Specific \& Related \& General Neurons}
    \end{minipage}
\caption{PPL changes of MistralMathOctopus on MSVAMP after deactivating language-specific neurons or language-specific \& language-related neurons or language-specific \& language-related \& general neurons.}
\label{fig:mistral-msvamp-heatmap-ppl}
\end{figure*}

\begin{figure*}[htbp]
    \centering
    \begin{minipage}[b]{0.33\textwidth}
    \includegraphics[width=1.0\linewidth]{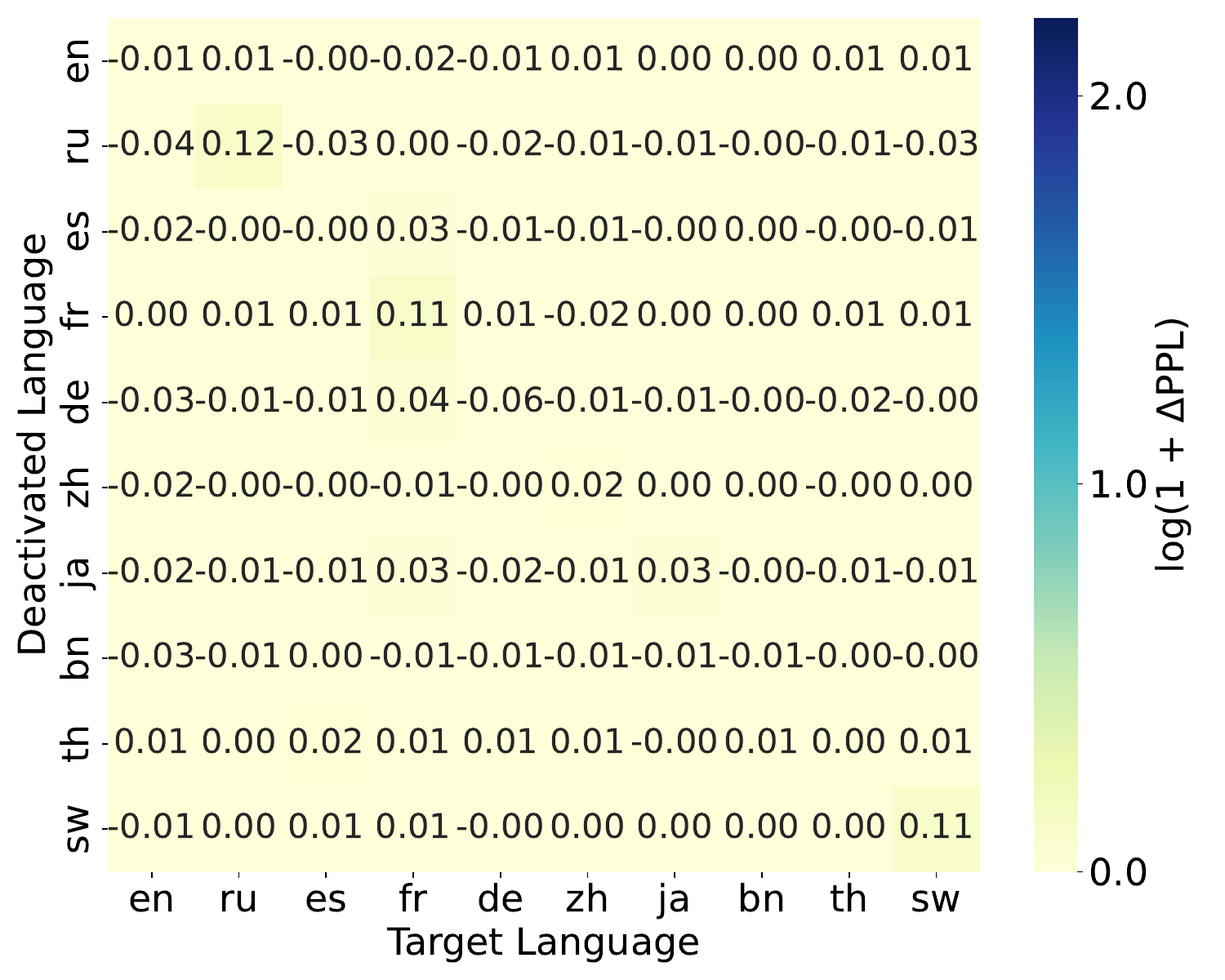}
    \subcaption{Specific Neurons}
    \end{minipage}
    \begin{minipage}[b]{0.33\textwidth}
    \includegraphics[width=1.0\linewidth]{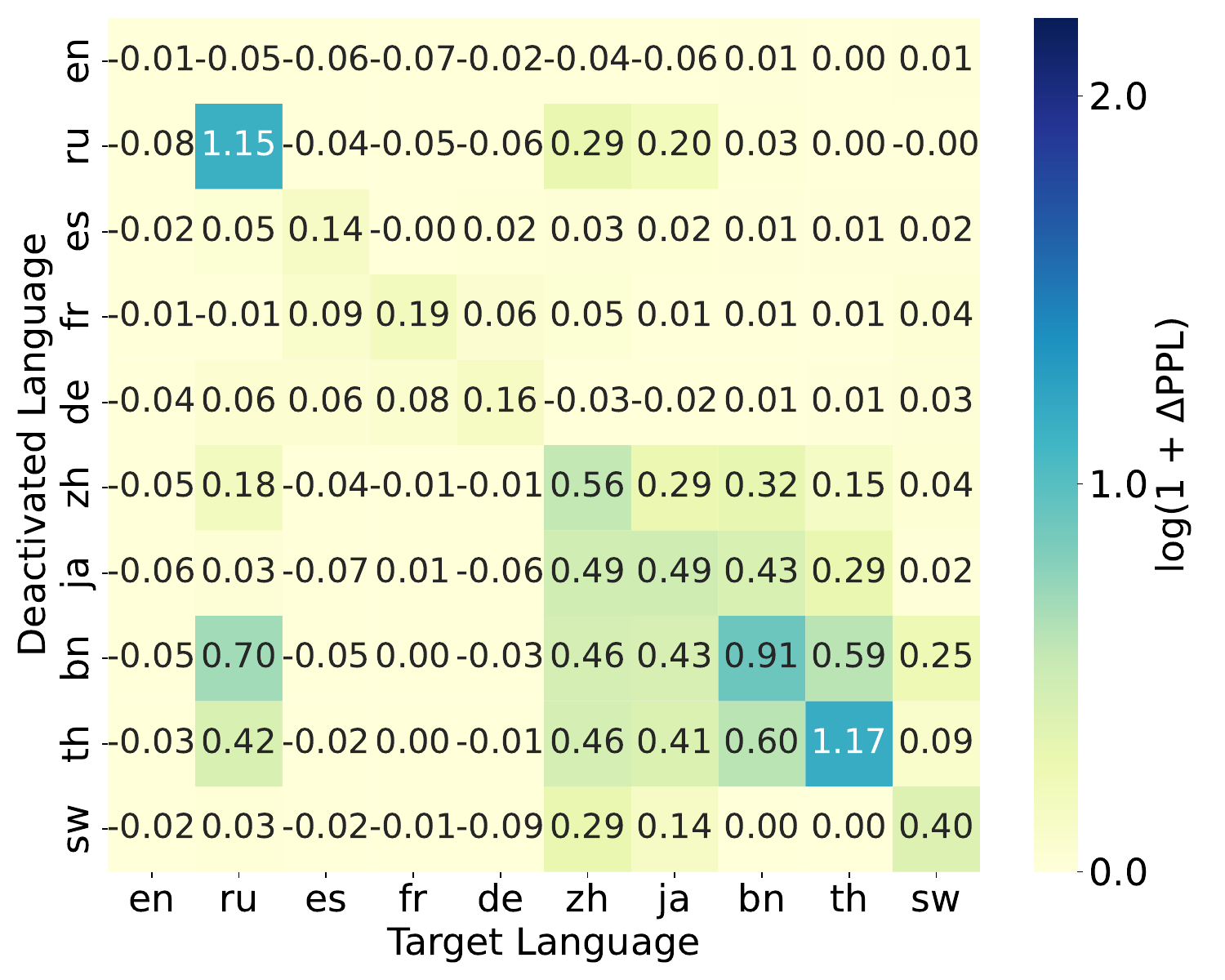}
    \subcaption{Specific \& Related Neurons}
    \end{minipage}
    \begin{minipage}[b]{0.33\textwidth}
    \includegraphics[width=1.0\linewidth]{Images/202602-Plots/ppl-Llama-mgsm-general.pdf}
    \subcaption{Specific \& Related \& General Neurons}
    \end{minipage}
\caption{PPL changes of MetaMathOctopus on MGSM after deactivating language-specific neurons or language-specific \& language-related neurons or language-specific \& language-related \& general neurons.}
\label{fig:llama-mgsm-heatmap-ppl}
\end{figure*}

\begin{figure*}[htbp]
    \centering
    \begin{minipage}[b]{0.33\textwidth}
    \includegraphics[width=1.0\linewidth]{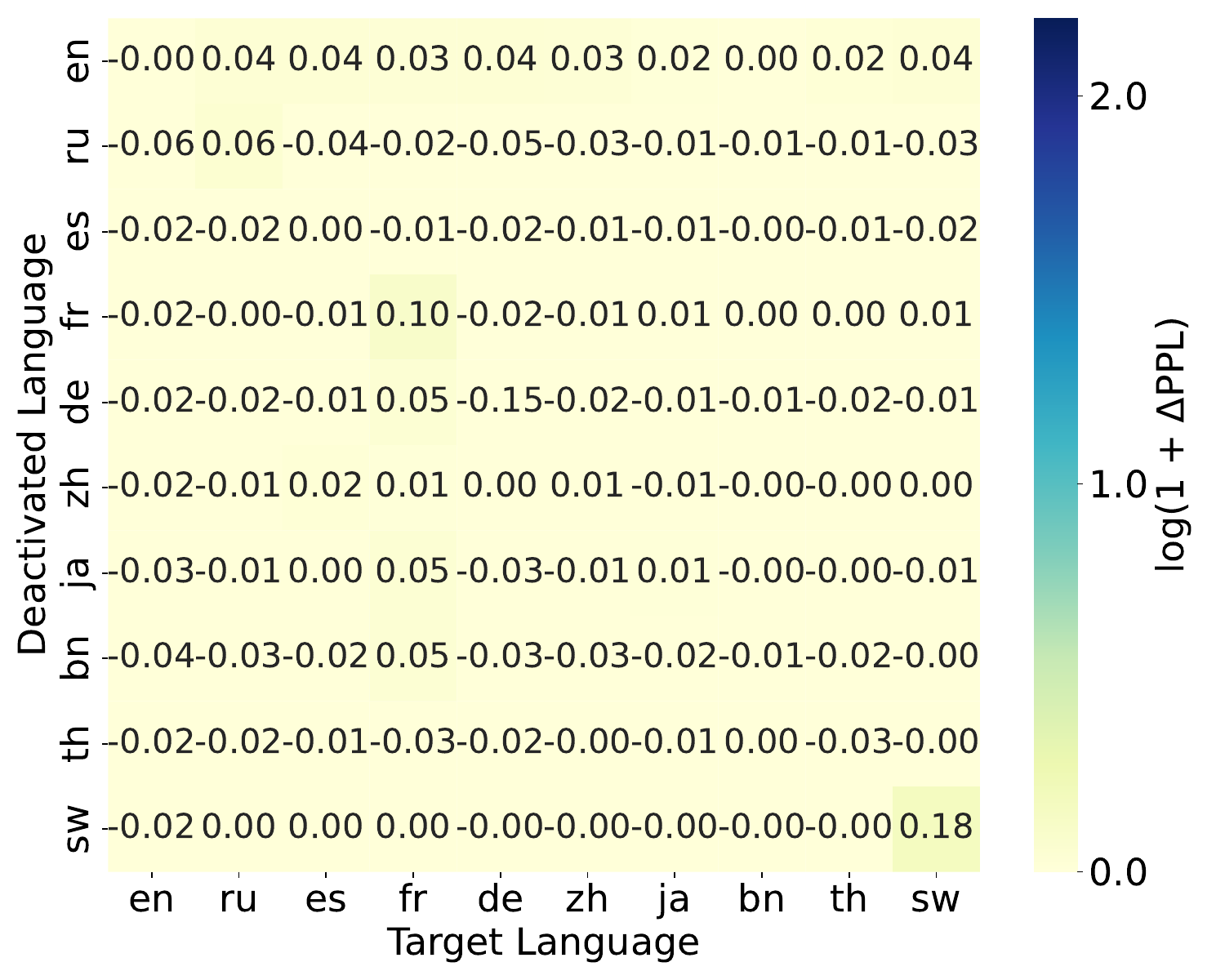}
    \subcaption{Specific Neurons}
    \end{minipage}
    \begin{minipage}[b]{0.33\textwidth}
    \includegraphics[width=1.0\linewidth]{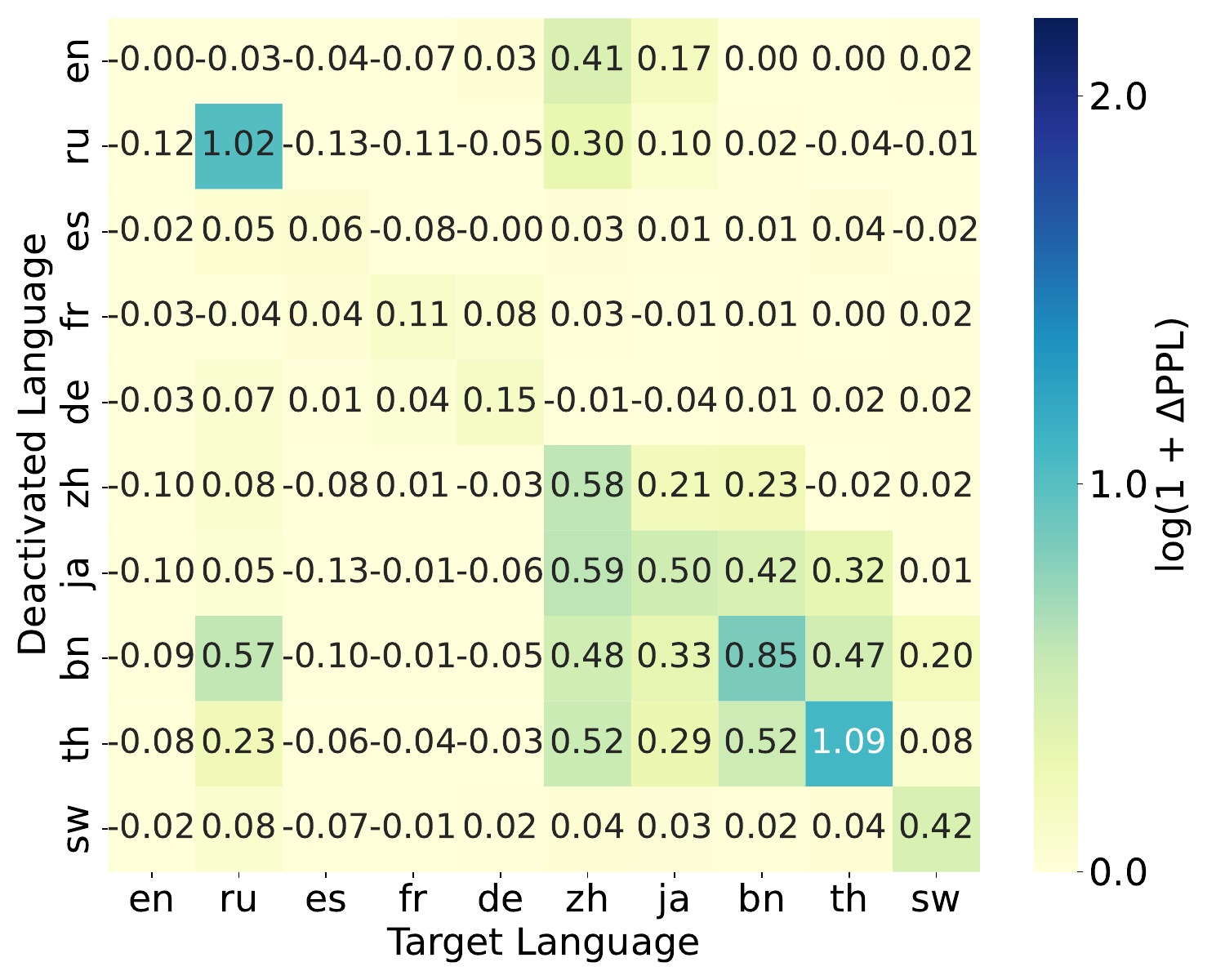}
    \subcaption{Specific \& Related Neurons}
    \end{minipage}
    \begin{minipage}[b]{0.33\textwidth}
    \includegraphics[width=1.0\linewidth]{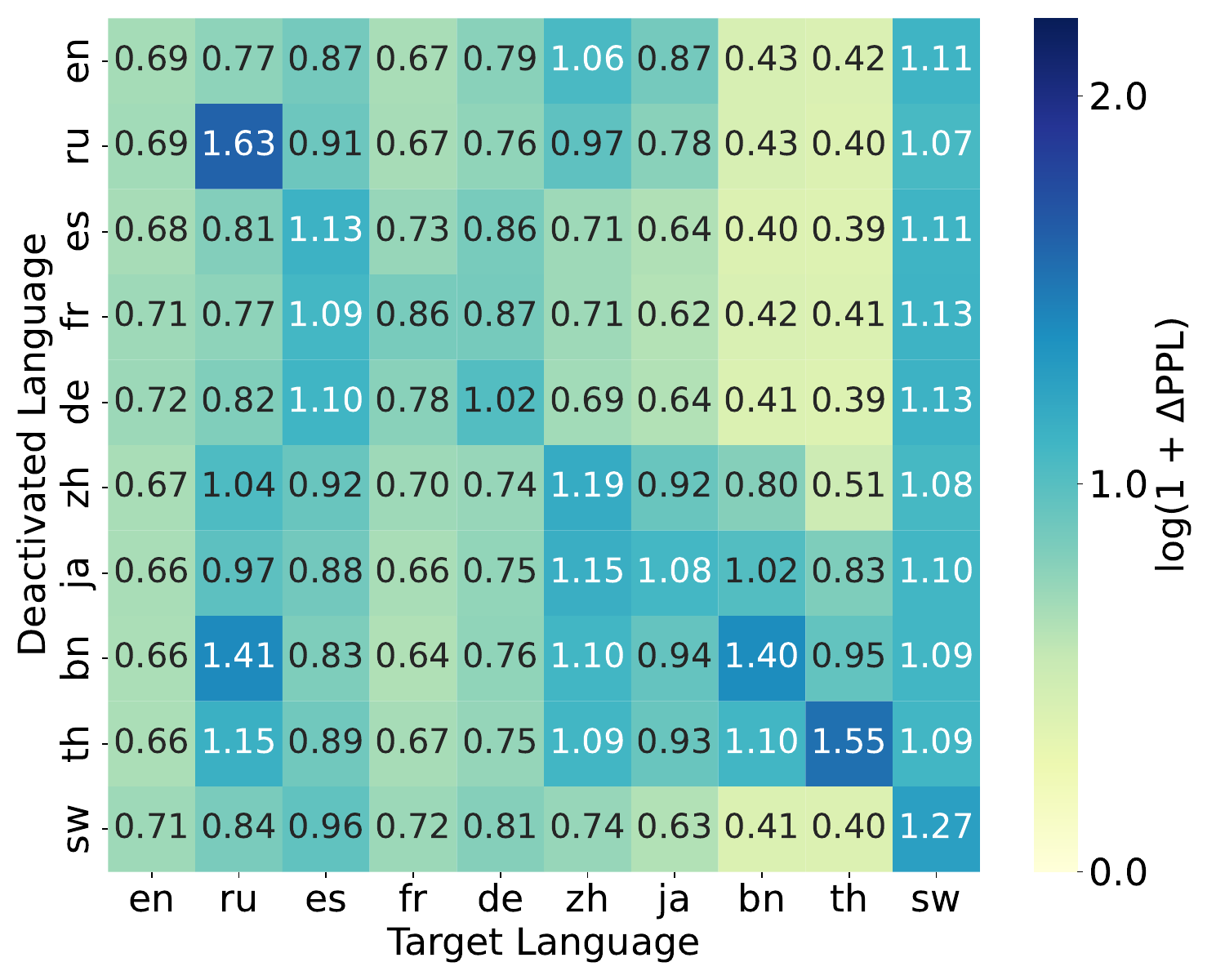}
    \subcaption{Specific \& Related \& General Neurons}
    \end{minipage}
\caption{PPL changes of MetaMathOctopus on MSVAMP after deactivating language-specific neurons or language-specific \& language-related neurons or language-specific \& language-related \& general neurons.}
\label{fig:llama-msvamp-heatmap-ppl}
\end{figure*}

For the MistralMathOctopus model, Figure~\ref{fig:baseline-heatmap} presents the results of the deactivation ablation experiments conducted on MGSM using the neuron identification algorithm proposed by~\citet{tang-etal-2024-language}. Additionally, Figure~\ref{fig:mistral-msvamp-heatmap-acc} presents the results of the deactivation ablation experiments on the out-of-domain test set, MSVAMP.
To verify the generalization of our identification algorithm, we also deploy the MetaMathOctopus to conduct the deactivation ablation experiment, with the result in 
Figure~\ref{fig:llama-mgsm-heatmap-acc} and Figure~\ref{fig:llama-msvamp-heatmap-acc}. Furthermore, following~\citet{tang-etal-2024-language}, we also report the PPL results of the above models and datasets in Figure~\ref{fig:mistral-mgsm-heatmap-ppl},~\ref{fig:mistral-msvamp-heatmap-ppl},~\ref{fig:llama-mgsm-heatmap-ppl}, and~\ref{fig:llama-msvamp-heatmap-ppl}.

\section{Layer-wise distribution of the different types of neurons}
\begin{figure*}[htbp]
    \centering 

    \begin{subfigure}[b]{0.47\textwidth}
    \includegraphics[width=0.9\linewidth]{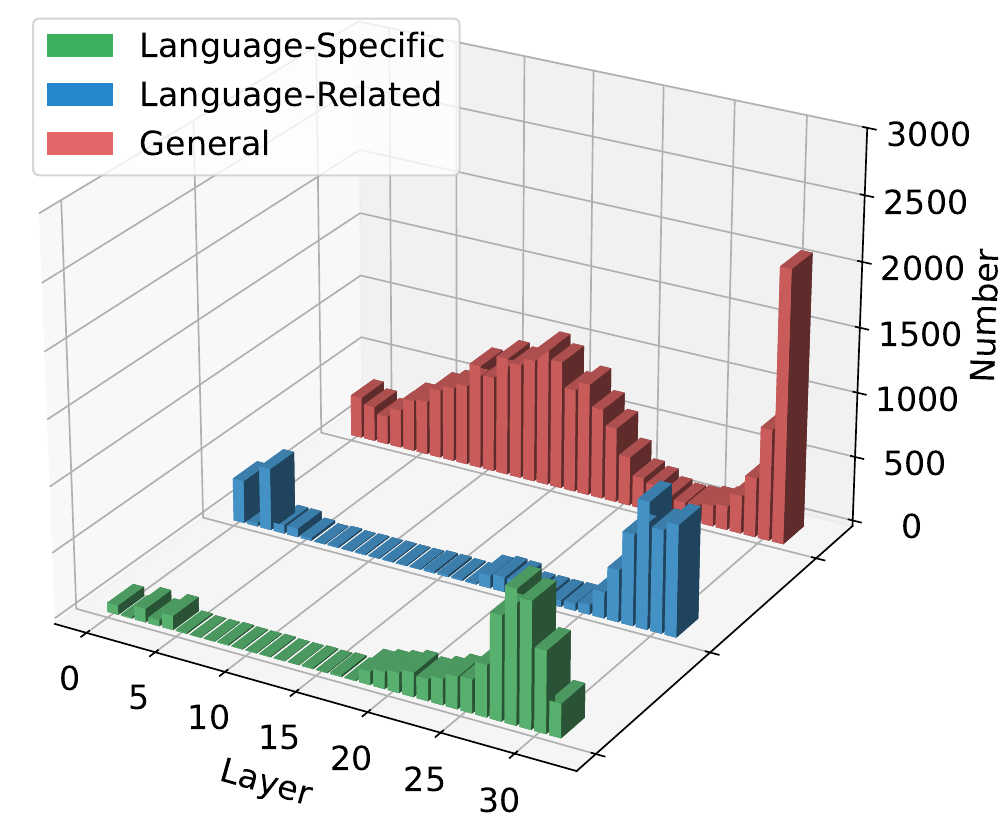}
    \caption{MistralMathOctopus on MSVAMP.}
    \label{fig:layer-wise-neurons-distribution_MSVAMP}
    \end{subfigure}
        \hspace{10pt}
    \begin{subfigure}[b]{0.47\textwidth}
    \includegraphics[width=0.9\linewidth]{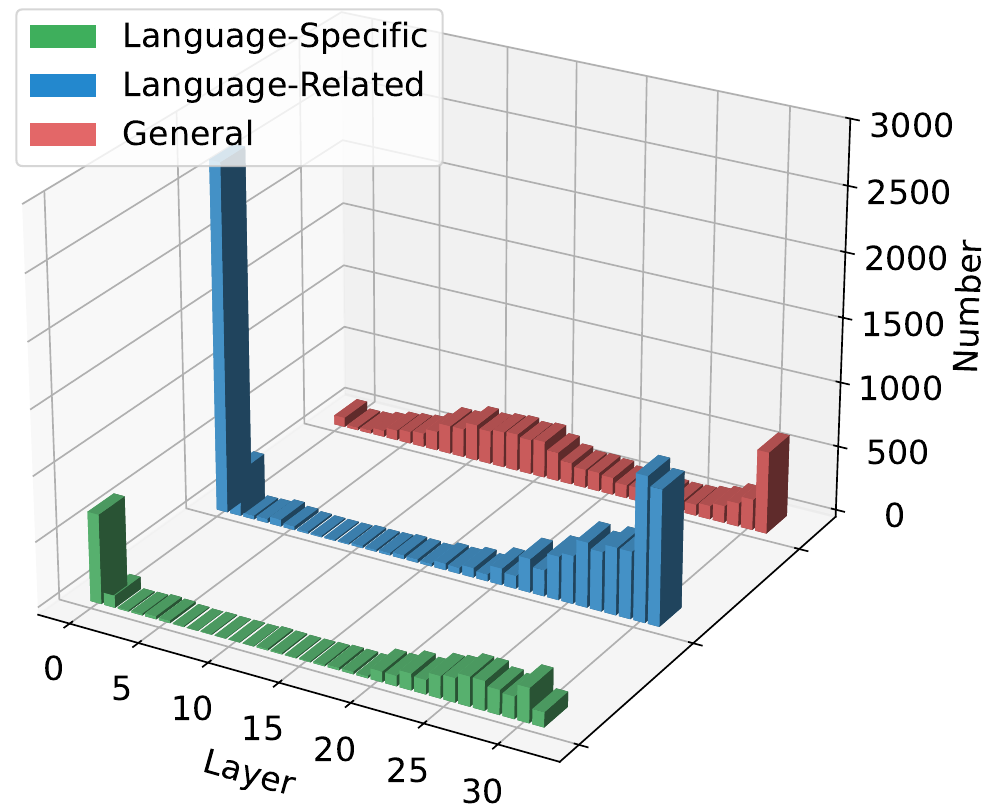}
    \caption{MetaMathOctopus on MGSM.}
    \label{fig:layer-wise-neurons-distribution_MGSM_llama}
    \end{subfigure}
    \caption{Layer-wise distribution of the different types of neurons.}
\end{figure*}

In Figure~\ref{fig:layer-wise-neurons-distribution_MSVAMP} and~\ref{fig:layer-wise-neurons-distribution_MGSM_llama}, we separately display the layer-wise distribution of the different types of neurons of MistarlMathOctopus on MSVAMP and MetaMathOctopus on MGSM to validate the generalization of our findings.

\section{Layer-wise Changes in the Number of Different Types of Neurons During Alignment}\label{app:checkpoint-layer-wise-changes}
\begin{figure*}[htbp]
    \centering 
    \includegraphics[width=0.52\textwidth]{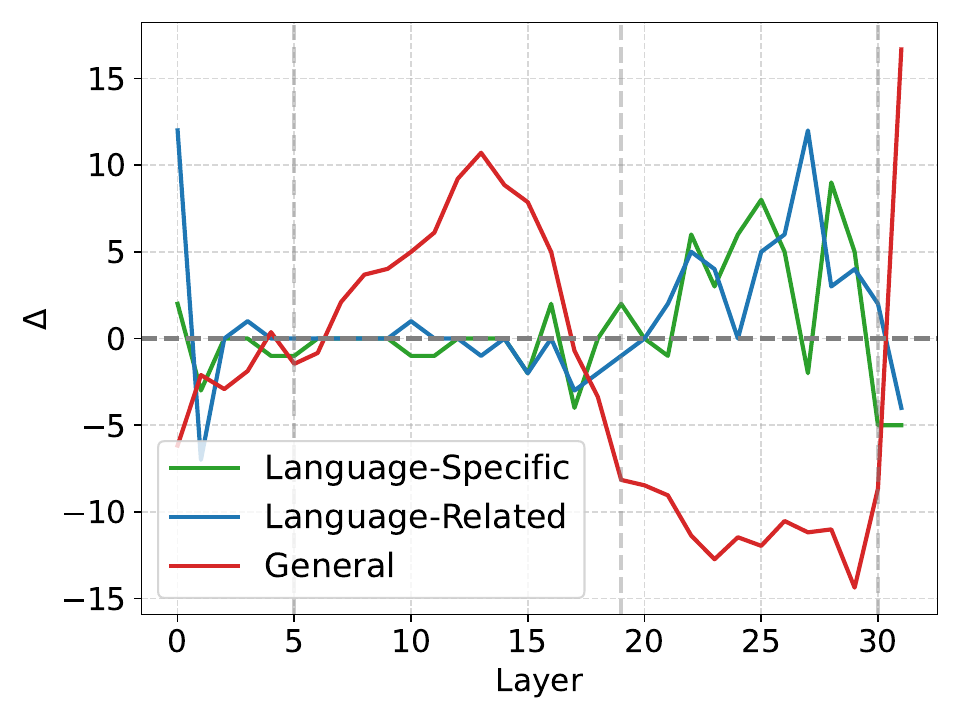}
    \caption{Layer-wise changes in the number of MetaMathOctopus}
    \label{fig:Layer-wise_changes_llama}
\end{figure*}

\begin{figure*}[htbp]
    \centering
    \begin{subfigure}[b]{0.45\textwidth}
    \includegraphics[width=\linewidth]{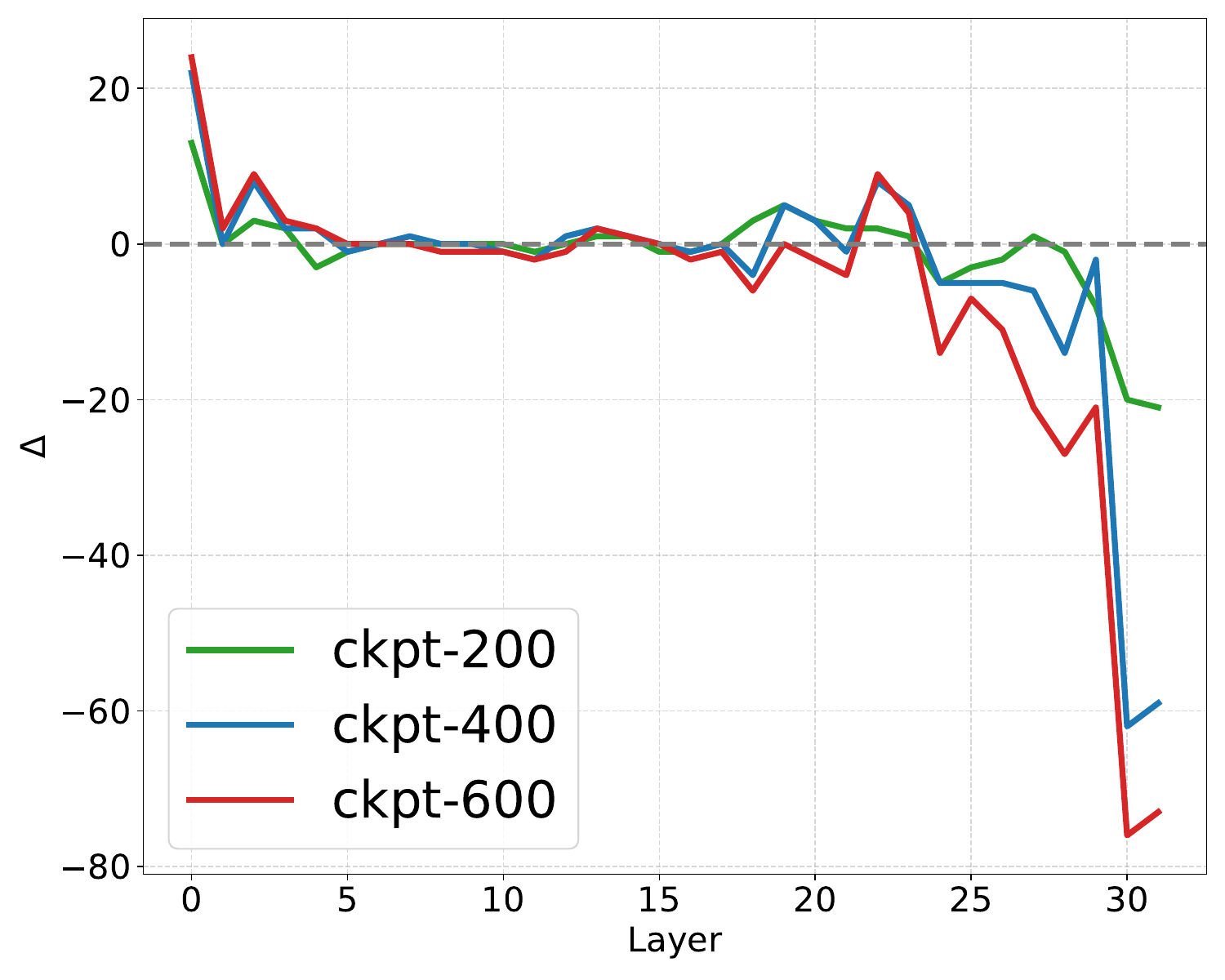}
    \caption{MGSM}
    \end{subfigure}
    \hspace{10pt}
    \begin{subfigure}[b]{0.45\textwidth}
    \includegraphics[width=\linewidth]{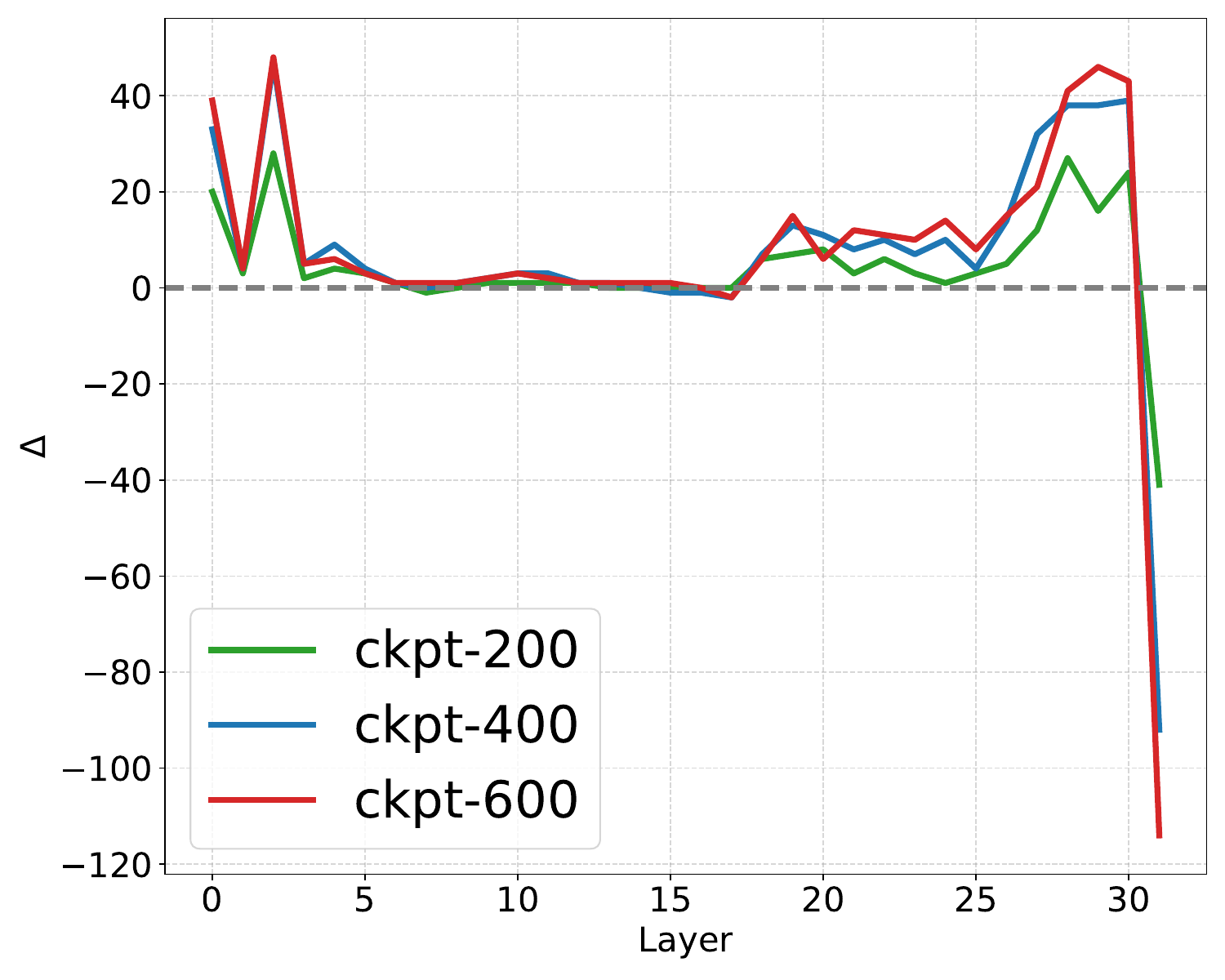}
    \caption{MSVAMP}
    \end{subfigure}
    \vspace{0.5em}
    \caption{Layer-wise changes in the number of language-specific neurons of MistralMathOctopus during the alignment.}
    \label{fig:Layer-wise_changes_specific}
\end{figure*}

\begin{figure*}[htbp]
    \centering 
    \begin{subfigure}[b]{0.45\textwidth}
    \includegraphics[width=\linewidth]{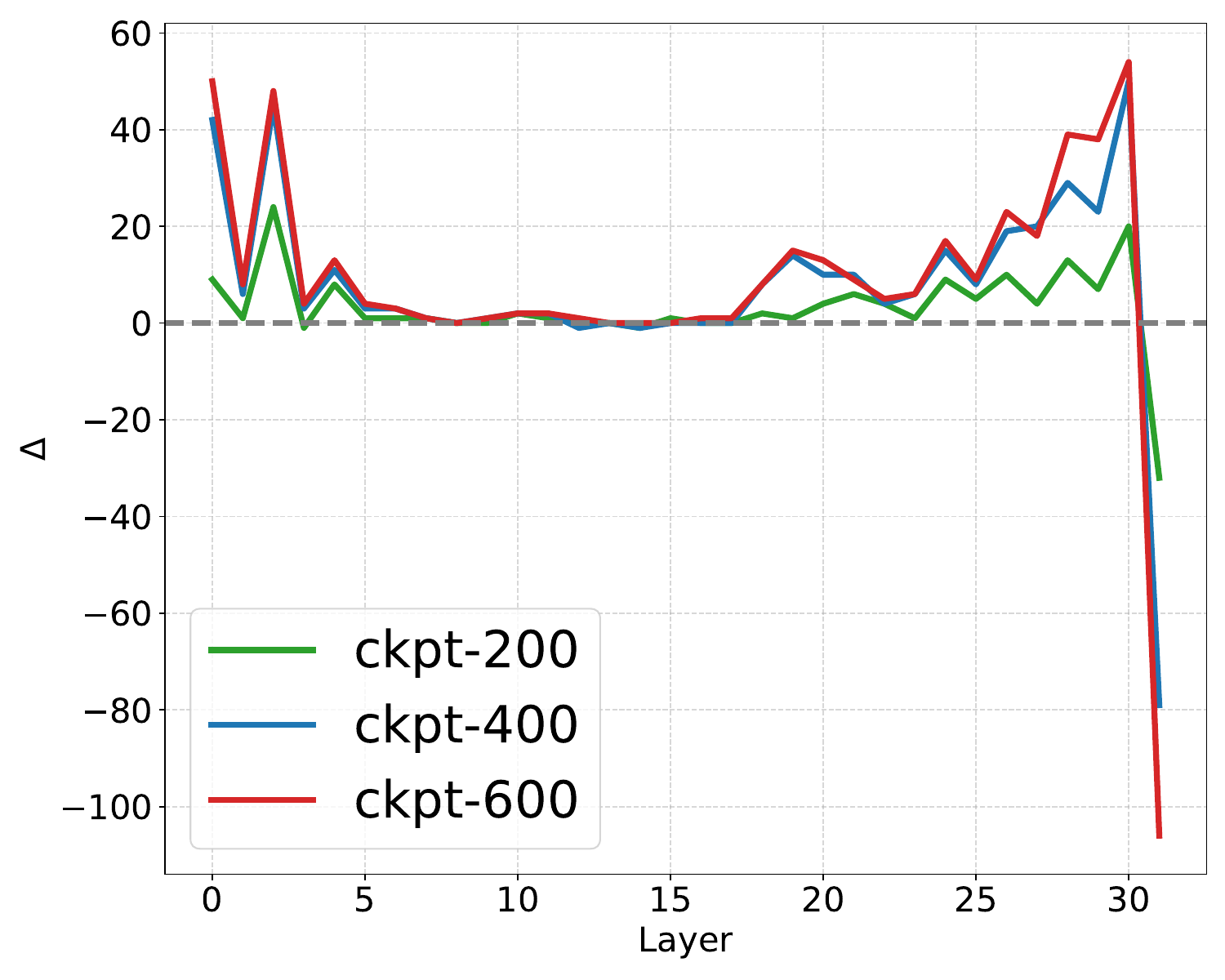}
    \caption{MGSM}
    \end{subfigure}
    \hspace{10pt}
    \begin{subfigure}[b]{0.45\textwidth}
    \includegraphics[width=\linewidth]{Images/layer_wise_delta_related_neurons_mistral_MSVAMP.pdf}
    \caption{MSVAMP}
    \end{subfigure}
    \vspace{0.5em}
    \caption{Layer-wise changes in the number of language-related neurons of MistralMathOctopus during the alignment.}
    \label{fig:Layer-wise_changes_related}
\end{figure*}

\begin{figure*}[htbp]
    \centering 
    \begin{subfigure}[b]{0.45\textwidth}
    \includegraphics[width=\linewidth]{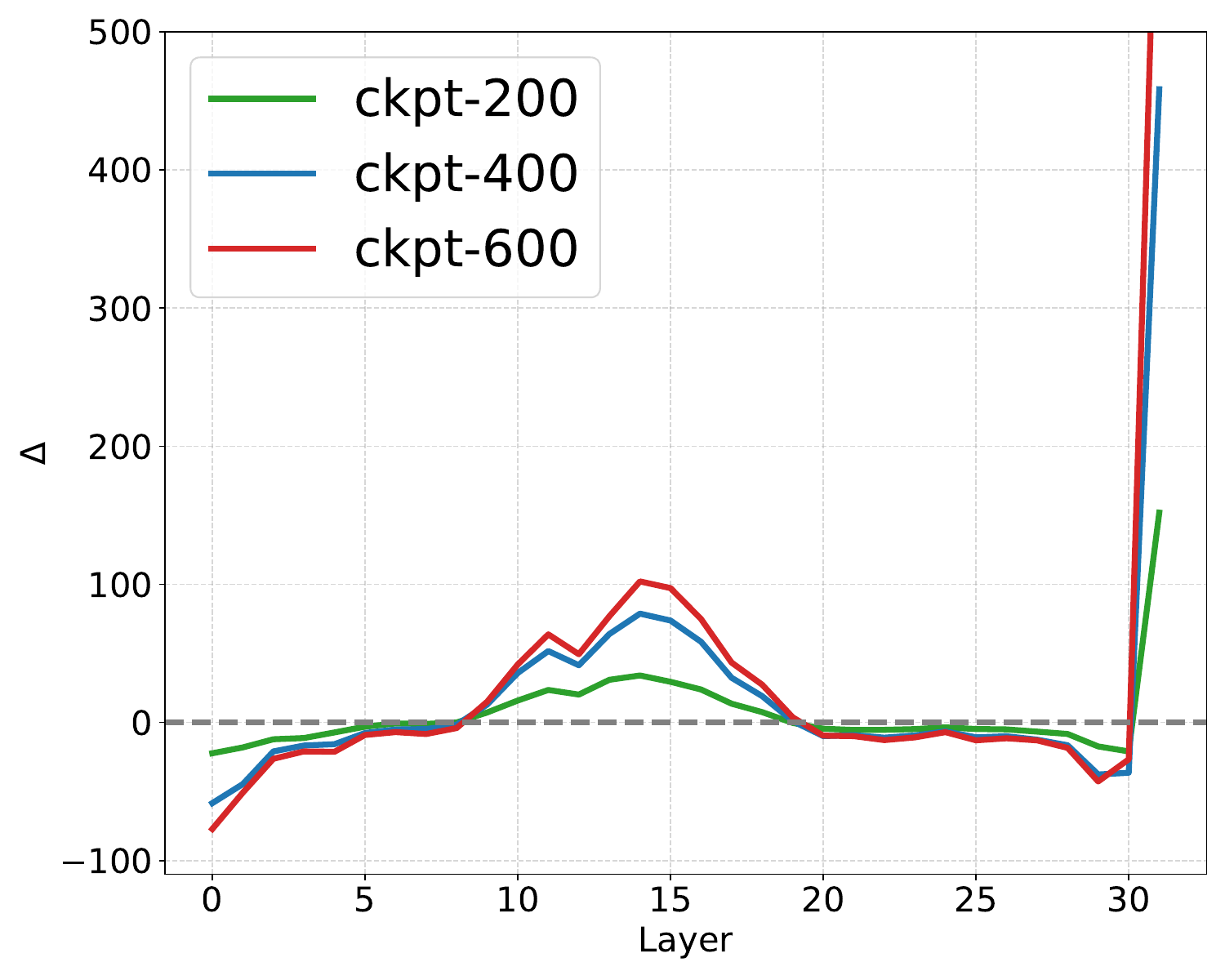}
    \caption{MGSM}
    \end{subfigure}
    \hspace{10pt}
    \begin{subfigure}[b]{0.45\textwidth}
    \includegraphics[width=\linewidth]{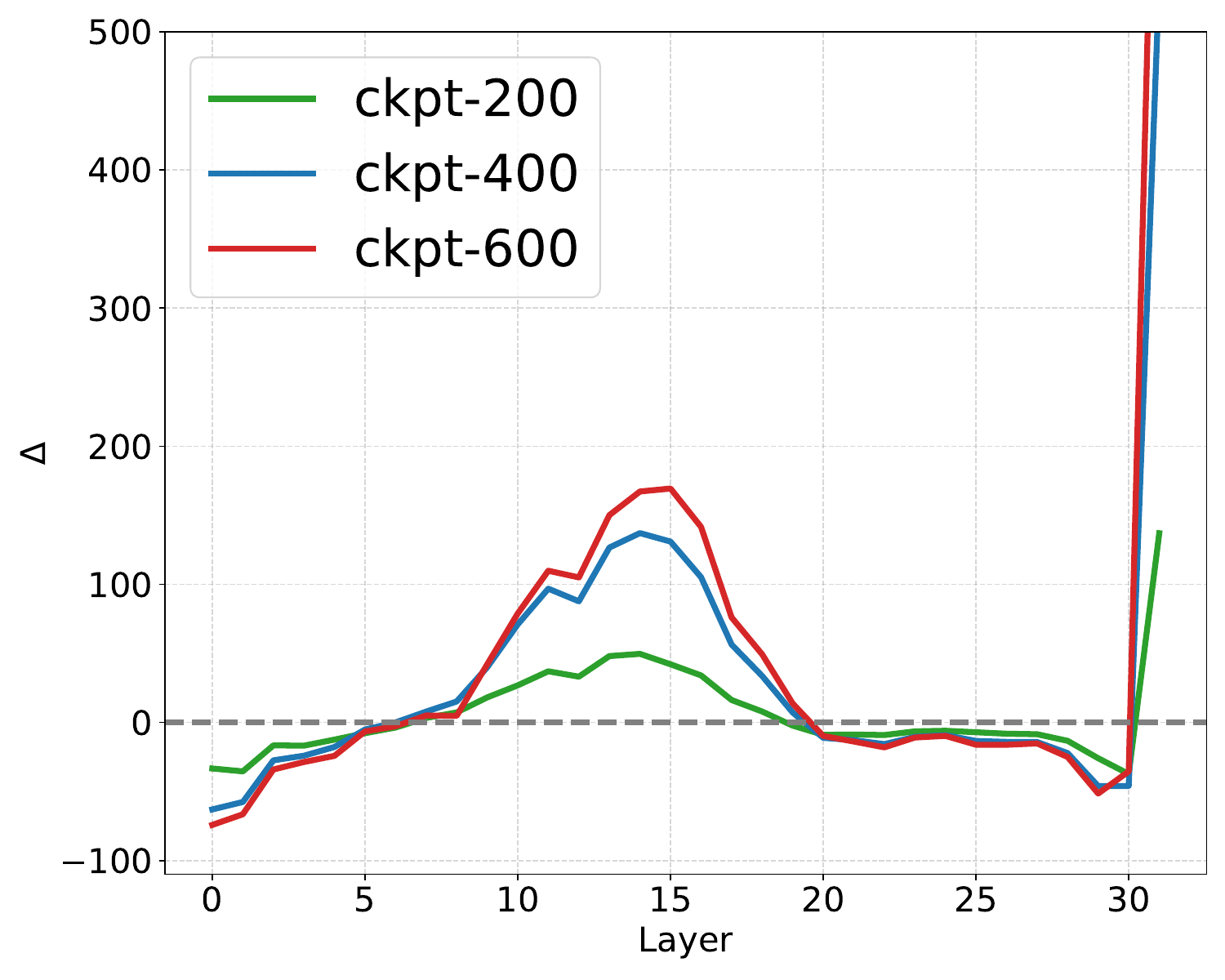}
    \caption{MSVAMP}
    \end{subfigure}
    \vspace{0.5em}
    \caption{Layer-wise changes in the number of general neurons of MistralMathOctopus during the alignment.}
    \label{fig:Layer-wise_changes_agnostic}
\end{figure*}

Figure~\ref{fig:Layer-wise_changes_llama} shows the same pattern as the results in the main text, thereby supporting its generalization.
Furthermore, for each dataset, we examine the layer-wise changes in the number of different types of neurons in the process of alignment in Figure~\ref{fig:Layer-wise_changes_specific},~\ref{fig:Layer-wise_changes_related} and~\ref{fig:Layer-wise_changes_agnostic}. The notation “$\text{ckpt-}x$” denotes the model checkpoint obtained after $x$ training steps.

\section{Spontaneous Multilingual Alignment}
We report the accuracy of the base model on the MGSM and MSVAMP benchmarks in Table~\ref{tab:spontaneous_result_all_mistral}. The expected accuracy of random guessing is 50.0\%. "X/Y $\Rightarrow$ T" indicates that languages X and Y are used for alignment. The best results for each language are highlighted. It is evident that models trained on multilingual translation data substantially outperform the original model across a wide range of languages, indicating that multilingual training significantly enhances the model's multilingual capabilities.

\section{Limitations}\label{app:limitations}
We provide insights and analysis results for multilingual alignment and multilingual capabilities of LLMs, which allows for a better understanding of multilingualism. Despite we have conducted a systematic investigation in our work, there are still some limitations waiting for research. Due to the limited resources, we only conduct experiments on two different models and two different datasets. We are willing to perform a more comprehensive analysis on different scenarios if more resources are available in the future. Additionally, we don't perform a finer-grained analysis of neurons within the same layer in this work. We would like to explore this in our future work.

\newpage

\begin{figure*}[htbp]
    \centering
    \begin{subfigure}[b]{0.45\textwidth}
    \includegraphics[width=\linewidth]{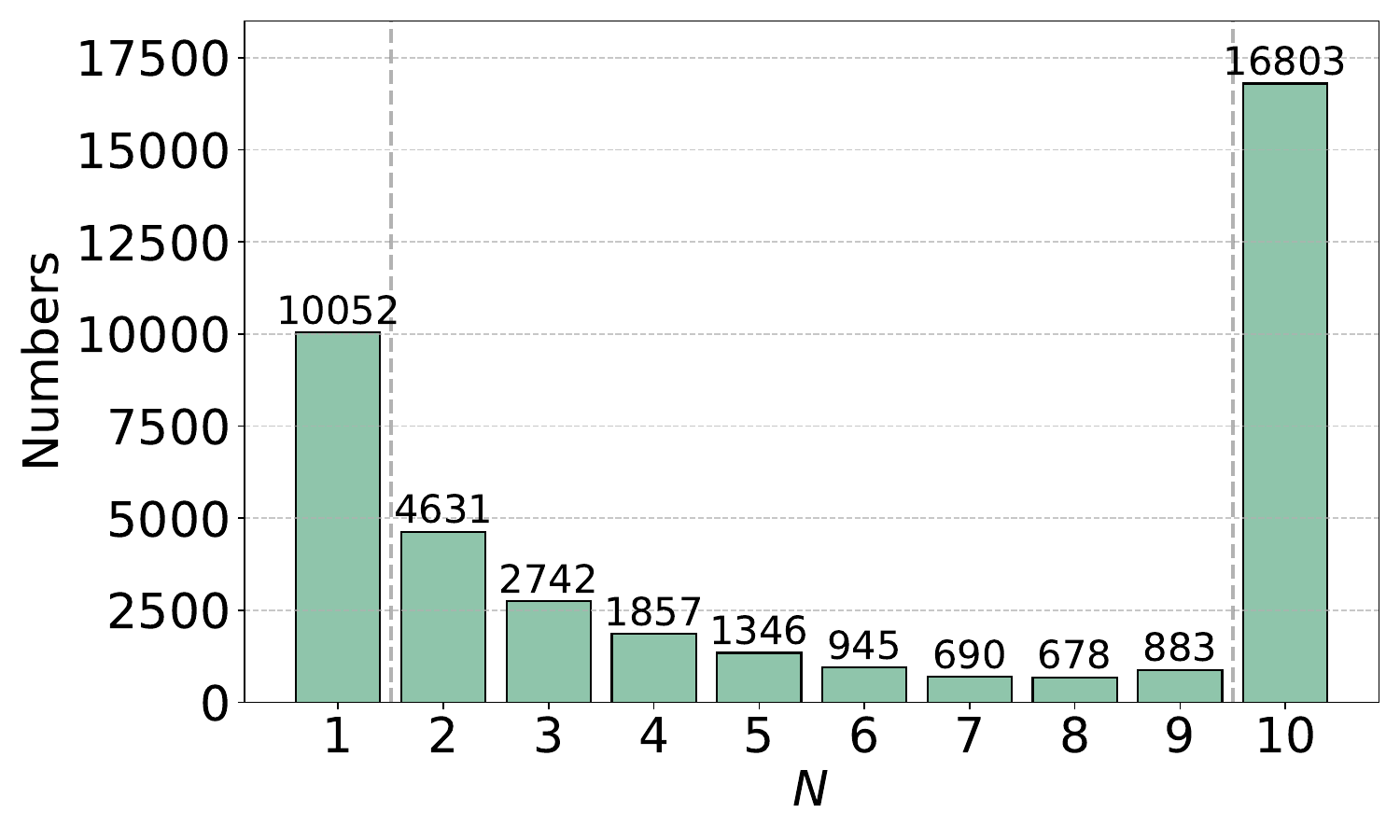}
    \caption{Before alignment}
    \end{subfigure}
    \hspace{10pt}
    \begin{subfigure}[b]{0.45\textwidth}
    \includegraphics[width=\linewidth]{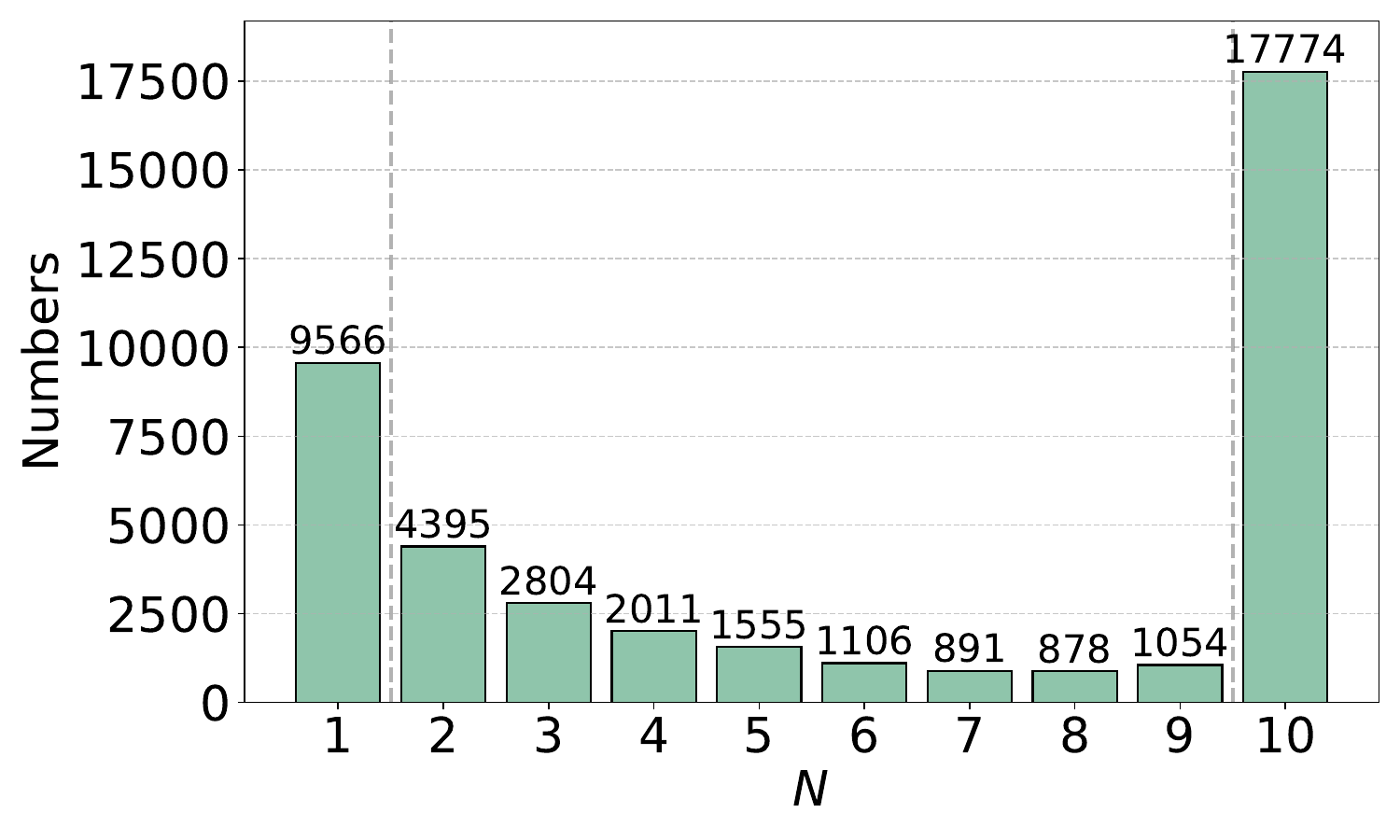}
    \caption{After alignment}
    \end{subfigure}
    \vspace{0.5em}
    \caption{Numbers of neurons shared by $N$ languages before and after alignment.}
    \label{fig:Related-Neuron_Numbers}
\end{figure*}

\begin{table*}[htbp]
\centering
\setlength{\tabcolsep}{4pt}
\renewcommand{\arraystretch}{1.2}
\resizebox{0.8\textwidth}{!}{
\begin{tabular}{l|cccccccccc|c}
\toprule
\textbf{Tested on  MGSM} & \textbf{bn} & \textbf{th} & \textbf{sw} & \textbf{ja} & \textbf{zh} & \textbf{ru} & \textbf{de} & \textbf{es} & \textbf{fr} & \textbf{en} & \textbf{Avg.} \\
\midrule
base & 43.6 & 53.2 & 50.4 & 55.6 & 59.6 & 59.2 & 61.2 & 62.8 & 56.8 & 75.6 & 57.8 \\
zh $\Rightarrow$ en & 49.6 & 58.4 & 54.8 & 56.4 & 65.2 & 70.0 & 66.4 & 72.8 & 68.8 & 78.4 & 64.1 \\
zh/de $\Rightarrow$ en & 46.4 & 55.6 & 59.2 & 56.8 & 64.0 & 71.2 & 66.8 & 71.2 & 69.2 & 75.2 & 63.6 \\
sw/th $\Rightarrow$ en & 48.8 & 58.8 & 59.2 & 56.4 & 68.4 & 68.4 & 69.2 & 69.6 & 70.4 & 77.6 & 64.7 \\
zh/es/ru  $\Rightarrow$ en & 46.0 & 56.4 & 58.8 & 54.8 & 63.2 & 70.8 & 68.8 & 71.6 & 69.6 & 76.8 & 63.7 \\
zh/es/fr/ja/de/sw/ru/th/bn $\Rightarrow$ en & 49.6 & 60.0 & 56.4 & 56.4 & 64.4 & 64.8 & 65.2 & 65.2 & 61.2 & 76.4 & 62.0 \\
\bottomrule
\toprule
\textbf{Tested on  MSVAMP} & \textbf{bn} & \textbf{th} & \textbf{sw} & \textbf{ja} & \textbf{zh} & \textbf{ru} & \textbf{de} & \textbf{es} & \textbf{fr} & \textbf{en} & \textbf{Avg.} \\
\midrule
base & 49.3 & 62.5 & 60.6 & 60.9 & 67.4 & 64.9 & 66.5 & 67.6 & 67.2 & 77.0 & 64.4 \\
zh $\Rightarrow$ en & 52.8 & 62.5 & 60.9 & 63.7 & 66.6 & 67.5 & 69.4 & 69.8 & 69.7 & 76.5 & 65.9 \\
zh/de $\Rightarrow$ en & 53.2 & 62.0 & 62.9 & 65.3 & 66.1 & 68.7 & 69.9 & 69.0 & 70.1 & 77.0 & 66.4 \\
sw/th $\Rightarrow$ en & 53.8 & 67.4 & 65.1 & 67.7 & 71.9 & 71.4 & 71.9 & 72.0 & 72.8 & 78.9 & 69.3 \\
zh/es/ru $\Rightarrow$ en & 52.9 & 63.2 & 61.5 & 63.0 & 67.8 & 68.1 & 69.8 & 68.7 & 70.6 & 78.3 & 66.4 \\
zh/es/fr/ja/de/sw/ru/th/bn $\Rightarrow$ en & 54.9 & 65.9 & 65.6 & 68.6 & 69.7 & 70.2 & 72.3 & 71.6 & 71.6 & 77.3 & 68.8 \\
\bottomrule
\end{tabular}
}
\caption{Accuracy of the base model and aligned variants on benchmarks.}
\label{tab:spontaneous_result_all_mistral}
\end{table*}

\clearpage

\end{document}